%% file: main.tex
\pdfoutput=1
\documentclass[10pt,twocolumn,letterpaper]{article}

\usepackage[pagenumbers]{cvpr} 

\input{preamble}

\definecolor{cvprblue}{rgb}{0.21,0.49,0.74}
\usepackage[pagebackref,breaklinks,colorlinks,citecolor=cvprblue]{hyperref}

\def\confName{CVPR}
\def\confYear{2024}

\begin{document}
\title{
Cache Me if You Can:\\Accelerating Diffusion Models through Block Caching 
}

\author{
Felix Wimbauer$^{1,2,3}$\and
Bichen Wu$^{1}$\and
Edgar Schoenfeld$^{1}$\and
Xiaoliang Dai$^{1}$\and
Ji Hou$^{1}$\and
Zijian He$^{1}$\and
Artsiom Sanakoyeu$^{1}$\and
Peizhao Zhang$^{1}$\and
Sam Tsai$^{1}$\and
Jonas Kohler$^{1}$\and
Christian Rupprecht$^{4}$\and
Daniel Cremers$^{2,3}$\and
Peter Vajda$^{1}$\and Jialiang Wang$^{1}$\and
$^{1}$Meta GenAI \hspace{.5cm} $^{2}$Technical University of Munich \hspace{.5cm} $^{3}$MCML \hspace{.5cm} $^{4}$University of Oxford\\
{\tt\small felix.wimbauer@tum.de \hspace{.5cm} jialiangw@meta.com }
}

\makeatletter
\let\@oldmaketitle\@maketitle
\renewcommand{\@maketitle}{\@oldmaketitle
  \centering
\captionsetup{type=figure}
\includegraphics[width=\linewidth]{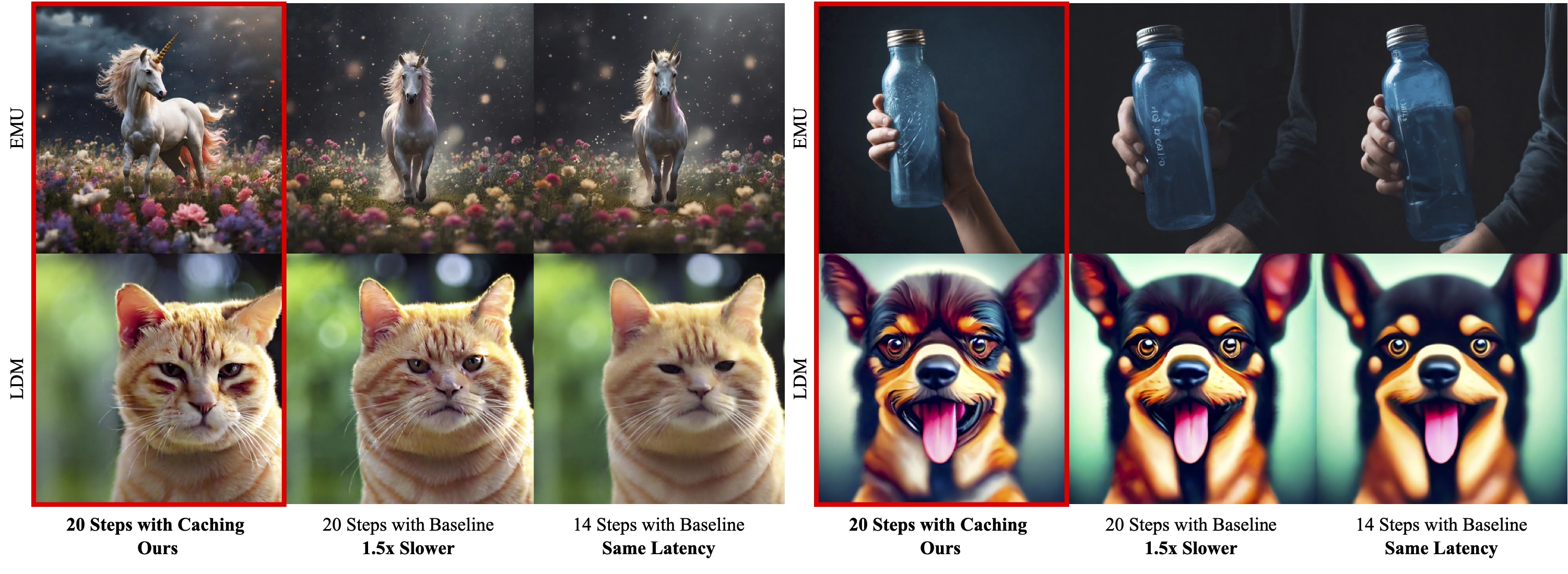}
\vspace{-.7cm}
\captionof{figure}{
\textbf{Speeding up diffusion models through block caching.} 
We observe that there are many redundant layer computations at different timesteps in diffusion models when generating an image.
Our block caching technique allows us to avoid these unnecessary computations, therefore speeding up inference by a factor of \textbf{1.5x}-\textbf{1.8x} while maintaining image quality.
Compared to the standard practice of naively reducing the number of denoising steps to match our inference speed, our approach produces more detailed and vibrant results. 
}
\vspace{.25cm}
  \label{fig:teaser}}

\maketitle
\let\thefootnote\relax\footnotetext{This work was done during Felix' internship at Meta GenAI.}
\input{sec/0_abstract}    
\input{sec/1_introduction}
\input{sec/2_related_work}

\input{sec/3_method}
\input{sec/4_experiments}
\input{sec/5_conclusion}

{
    \small
    \bibliographystyle{ieeenat_fullname}
    \bibliography{main}
}

\normalsize
\input{sec/X_suppl}

\end{document}

%% file: preamble.tex
\usepackage[dvipsnames]{xcolor}
\usepackage{graphicx}
\usepackage{amsmath}
\usepackage{amssymb}
\usepackage{booktabs}
\usepackage{mathtools}
\usepackage{multirow}
\usepackage{cuted}
\usepackage[dvipsnames]{xcolor}

\definecolor{armygreen}{rgb}{0.29, 0.43, 0.23} 

\definecolor{darkblue}{rgb}{0.11,0.29,0.54}


%% file: sec/0_abstract.tex
\begin{abstract}

Diffusion models have recently revolutionized the field of image synthesis due to their ability to generate photorealistic images.
However, one of the major drawbacks of diffusion models is that the image generation process is costly. 
A large image-to-image network has to be applied many times to iteratively refine an image from random noise.
While many recent works propose techniques to reduce the number of required steps, they generally treat the underlying denoising network as a black box.
In this work, we investigate the behavior of the layers \textit{within} the network and find that 
 1) the layers' output changes smoothly over time,
 2) the layers show distinct patterns of change, and 
 3) the change from step to step is often very small.
We hypothesize that many layer computations in the denoising network are redundant.
Leveraging this, we introduce block caching, in which we reuse outputs from layer blocks of previous steps to speed up inference.
Furthermore, we propose a technique to automatically determine caching schedules based on each block's changes over timesteps. 
In our experiments, we show through FID, human evaluation and qualitative analysis that Block Caching allows to generate images with higher visual quality at the same computational cost.
We demonstrate this for different state-of-the-art models (LDM and EMU) and solvers (DDIM and DPM).

\end{abstract}

%% file: sec/1_introduction.tex
\vspace{-.3cm}
\section{Introduction}
\label{sec:introduction}
Recent advances in diffusion models have revolutionized the field of 
generative AI. Such models are typically pretrained on billions of text-image pairs, and are commonly referred to as ``foundation models''. Text-to-image foundation models such as LDM~\cite{rombach2022high}, Dall-E 2/3~\cite{ramesh2022hierarchical, dalle3}, Imagen~\cite{saharia2022photorealistic}, and Emu~\cite{dai2023emu} can generate very high quality, photorealistic images that follow user prompts. These foundation models enable many downstream tasks, ranging from image editing \cite{brooks2022instructpix2pix, hertz2022prompt} to synthetic data generation 
\cite{hu2023gaia}, to video and 3D generations~\cite{singer2022make, poole2022dreamfusion}.

However, one of the drawbacks of such models is their high latency and computational cost. The denoising network, which typically is a U-Net with residual
and transformer blocks, tends to be very large in size and is repeatedly applied to obtain a final image. 
Such high latency prohibits many applications that require fast and frequent inferences. 
Faster inference makes large-scale image generation economically and technically viable.

The research community has made significant efforts to speed up image generation foundation models. 
Many works aim to reduce the number of steps required in the denoising process by changing the solver~\cite{lu2022dpm, lu2022dpm++, dockhorn2022genie, zhang2022fast,shaul2023bespoke}.
Other works propose to distill existing neural networks into architectures that require fewer steps~\cite{salimans2021progressive} or that can combine the conditional and unconditional inference steps~\cite{meng2023distillation}.  
While improved solvers and distillation techniques show promising results, 
they typically treat the U-Net model itself as a black box and mainly consider what to do with the network's output. 
This leaves a potential source of speed up---the U-Net itself---completely untapped.

In this paper, we investigate the denoising network in-depth, focusing on the behavior of attention blocks. Our observations reveal that:
1) The attention blocks change smoothly over denoising steps.
2) The attention blocks show distinct patterns of change depending on their position in the network. These patterns are different from each other, but they are consistent irrespective of the text inputs. 
3) The change from step to step is typically very small in the majority of steps. 
Attention blocks incur the biggest computational cost of most common denoising networks, making them a prime target to reduce network latency.

Based on these observations, we propose a technique called \textbf{block caching}. 
Our intuition is that if a layer block does not change much, we can avoid recomputing it 
to reduce redundant computations.
We extend this by a lightweight \textbf{scale-shift alignment mechanism}, which prevents artifacts caused by naive caching due to feature misalignment.
Finally, we propose an effective mechanism to \textbf{automatically derive caching schedules}. 

We analyse two different models: a retrained version of Latent Diffusion Models~\cite{rombach2022high} on Shutterstock data, 
as well as the recently proposed EMU~\cite{dai2023emu}, as can be seen in \cref{fig:teaser}. 
For both, we conduct experiments with two popular solvers: DDIM \cite{song2020denoising} and DPM \cite{lu2022dpm}.
For all combinations, given a fixed computational budget (inference latency), we can perform more steps with block caching and achieve better image quality.
Our approach achieves both improved FID scores and is preferred in independent human evaluations.

%% file: sec/2_related_work.tex
\section{Related Work}
In the following, we introduce important works that are related to our proposed method. 

\input{figures/analysis}

\paragraph{Text-to-Image Models.}
With recent advances in generative models, a vast number of text-conditioned models for image synthesis emerged.
Starting out with GAN-based methods \cite{goodfellow2014generative, zhang2017stackgan, zhang2018stackgan++, qiao2019mirrorgan, zhu2019dm, xu2018attngan, reed2016generative, li2019controllable, radford2015unsupervised, xia2021tedigan, tao2022df}, researchers discovered important techniques such as adding self-attention layers \cite{zhang2019self} for better long-range dependency modeling and scaling up to very large architectures \cite{brock2018large, kang2023scaling}. 
Different autoencoder-based methods \cite{razavi2019generating, he2022masked}, in particular generative transformers \cite{esser2021taming, chang2022maskgit, chen2020generative, ramesh2021zero}, can also synthesize new images in a single forward pass and achieve high visual quality. 
Recently, the field has been dominated by diffusion models \cite{sohl2015deep, song2020denoising, song2020score}. 
Advances such as classifier guidance \cite{dhariwal2021diffusion}, classifier-free guidance \cite{ho2021classifier, nichol2022glide}, and diffusion in the latent space \cite{rombach2022high} have enabled modern diffusion models \cite{dai2023emu, rombach2022high, nichol2022glide, ramesh2022hierarchical, chen2023pixart, balaji2022ediffi, feng2023ernie, xue2023raphael, saharia2022photorealistic} to generate photorealistic images at high resolution from text.
However, this superior performance often comes at a cost: 
Due to repeated applications of the underlying denoising neural network, image synthesis with diffusion models is very computationally expensive.
This not only hinders their widespread usage in end-user products, but also slows down further research.
To facilitate further democratization of diffusion models, we focus on accelerating diffusion models in this work.

\paragraph{Improved Solvers.}
In the diffusion model framework, we draw a new sample at every step from a distribution determined by the previous steps.
The exact sampling strategy, defined by the so-called solver, plays an important role in determining the number of steps we have to make to obtain high-quality output.
Starting out from the DDPM \cite{ho2020denoising} formulation, DDIM \cite{song2020denoising} introduced implicit probabilistic models.
DDIM allows the combination of DDPM steps without retraining and is popular with many current models.
The DPM-Solver \cite{lu2022dpm, lu2022dpm++} models the denoising process as an ordinary differential equation and proposes a dedicated high-order solver for diffusion ODEs.
Similar approaches are adopted by \cite{zhang2022fast, zhang2023improved, karras2022elucidating, zhao2023unipc, liu2022pseudo}.
Another line of works \cite{shaul2023bespoke, dockhorn2022genie, watson2021learning, lam2021bilateral, duan2023optimal} proposed to train certain parts of the solver on a dataset.
While better solvers can help to speed up image synthesis by reducing the number of required steps, they still treat the underlying neural network as a black box.
In contrast, our work investigates the internal behavior of the neural network and gains speed up from caching.
Therefore, the benefits of improved solvers and our caching strategy are not mutually exclusive. 

\paragraph{Distillation.}
Distillation techniques present an alternative way to speed up inference.
Here, a pretrained teacher network creates new training targets for a student architecture, that needs fewer neural function evaluations than the teacher.
Guidance distillation \cite{meng2023distillation} replaces the two function evaluations of classifier-free guidance with a single one, while progressive distillation \cite{salimans2021progressive} reduces the number of sampling steps. \cite{luhman2021knowledge} optimizes a student to directly predict the image generated by the teacher in one step.

Consistency models \cite{song2023consistency, luo2023latent} use a consistency formulation enabling a single-step student to do further steps.
Finally, \cite{yang2023diffusion} distill a large teacher model into a much smaller student architecture. However, distillation does not come without cost. Apart from the computational cost of re-training the student model, some distillation techniques cannot handle negative or composite prompts~\cite{meng2023distillation,liu2022compositional}. 
In this paper, we introduce a lightweight fine-tuning technique inspired by distillation, that leaves the original parameters unchanged while optimizing a small number of extra parameters without restricting the model.

%% file: figures/analysis.tex
\begin{figure*}[t]
\centering
\begin{tabular}{c c}
\begin{subfigure}[t]{0.54\textwidth}
    \centering 
    \hspace{-1.5em}
    \includegraphics[trim={0cm 0cm 0cm 0cm},clip,width=\textwidth]
    {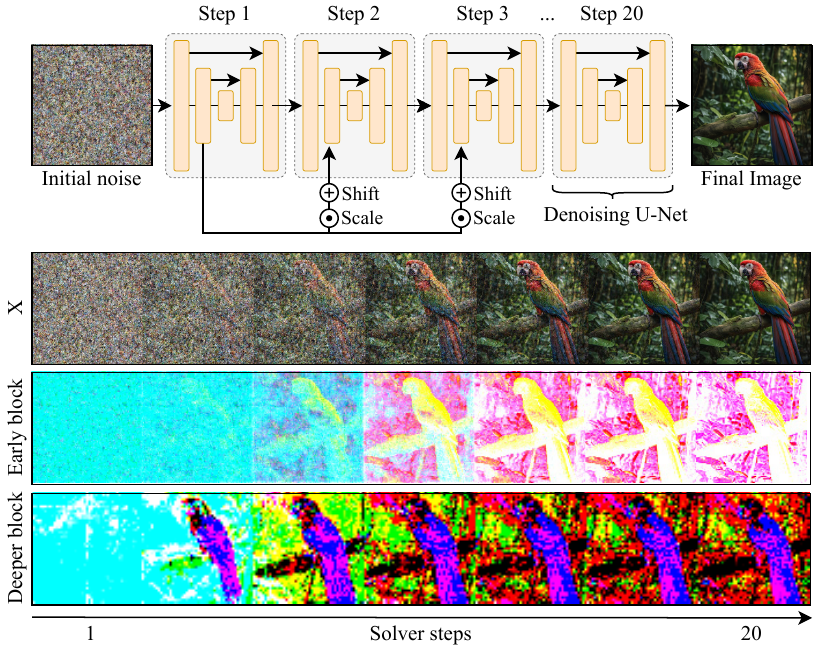} 
   %
\end{subfigure} & \begin{subfigure}[t]{0.44\textwidth}
    \centering
    \vspace{-21em} 
    \hspace{-2.5em}
     \setlength{\tabcolsep}{0em} 
    \renewcommand{\arraystretch}{0.3} 
    \begin{tabular}{c c}
    \includegraphics[trim={0.9cm 0cm 0cm 0cm},clip,height=0.47\textwidth ]{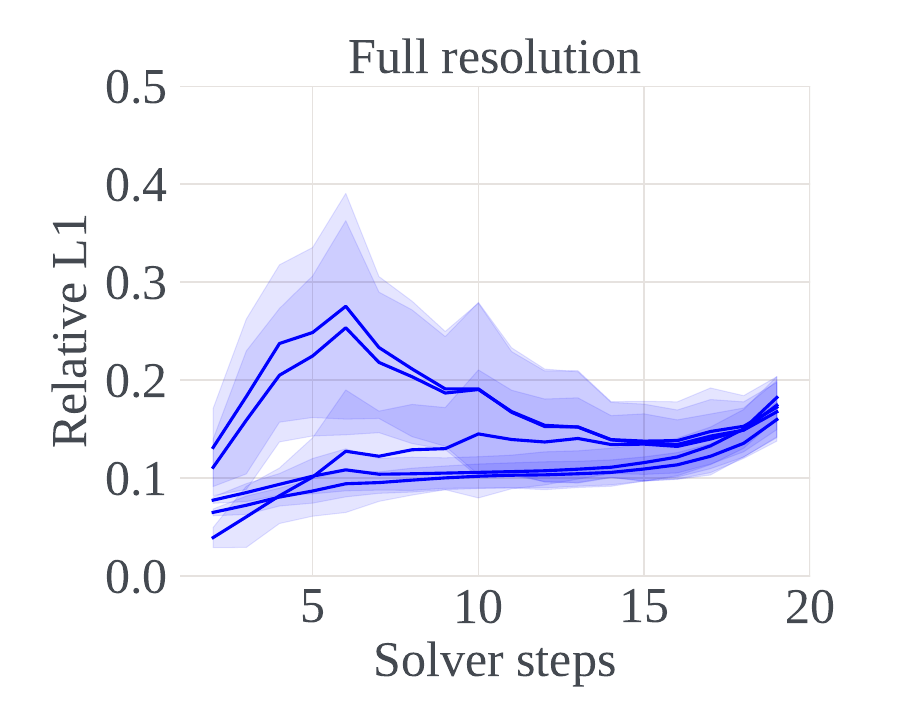} & 
    \includegraphics[trim={0.9cm 0cm 1cm 0cm},clip,height=0.47\textwidth]{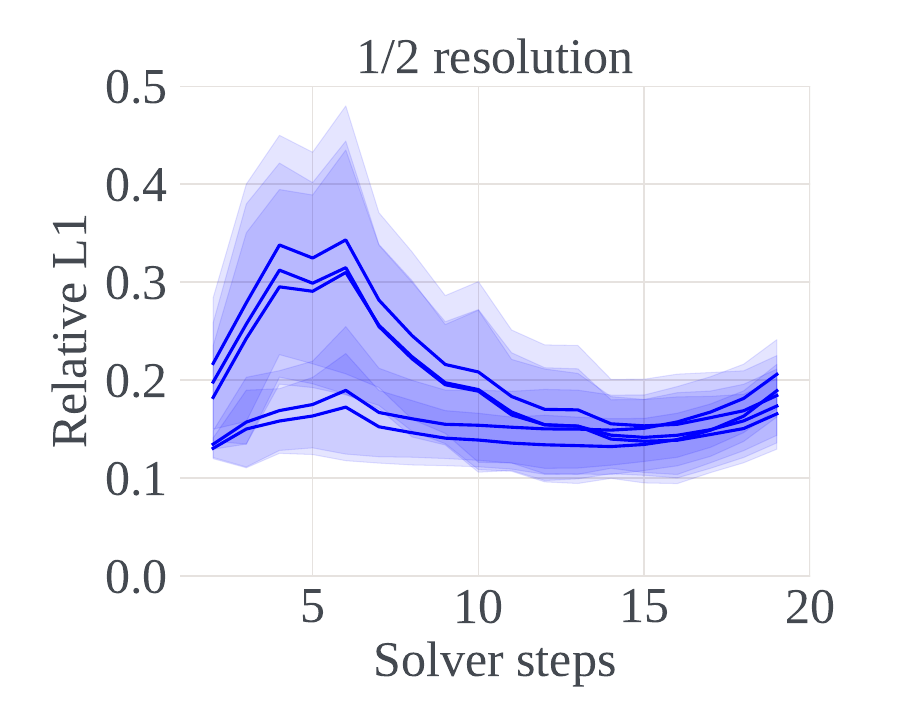}  \\
    \includegraphics[trim={0.9cm 0cm 1cm 0cm},clip,height=0.47\textwidth]{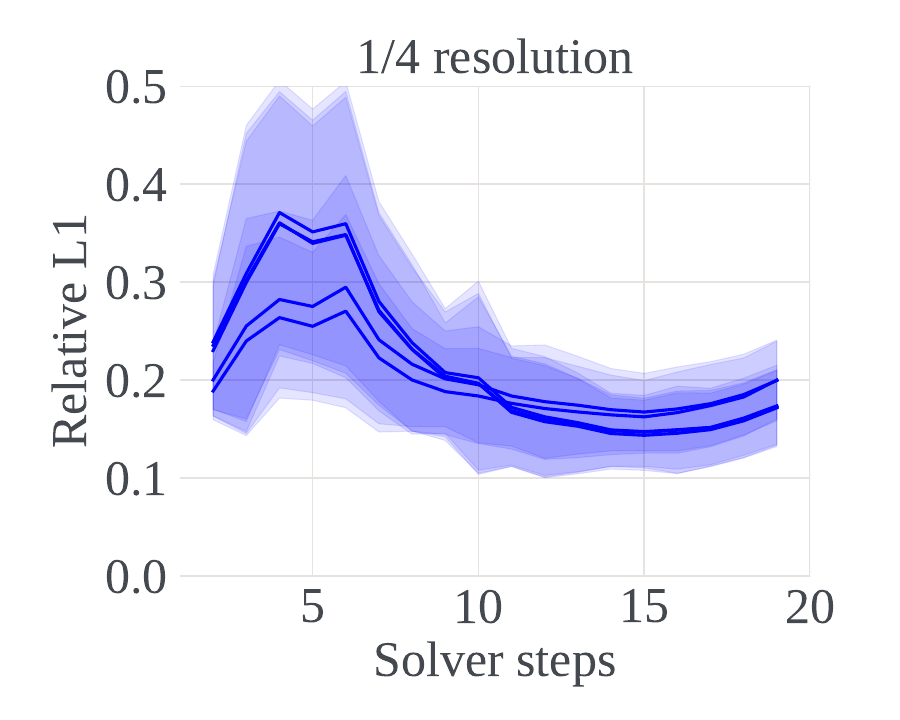} & 
    \includegraphics[trim={0.9cm 0cm 1cm 0cm},clip,height=0.47\textwidth]{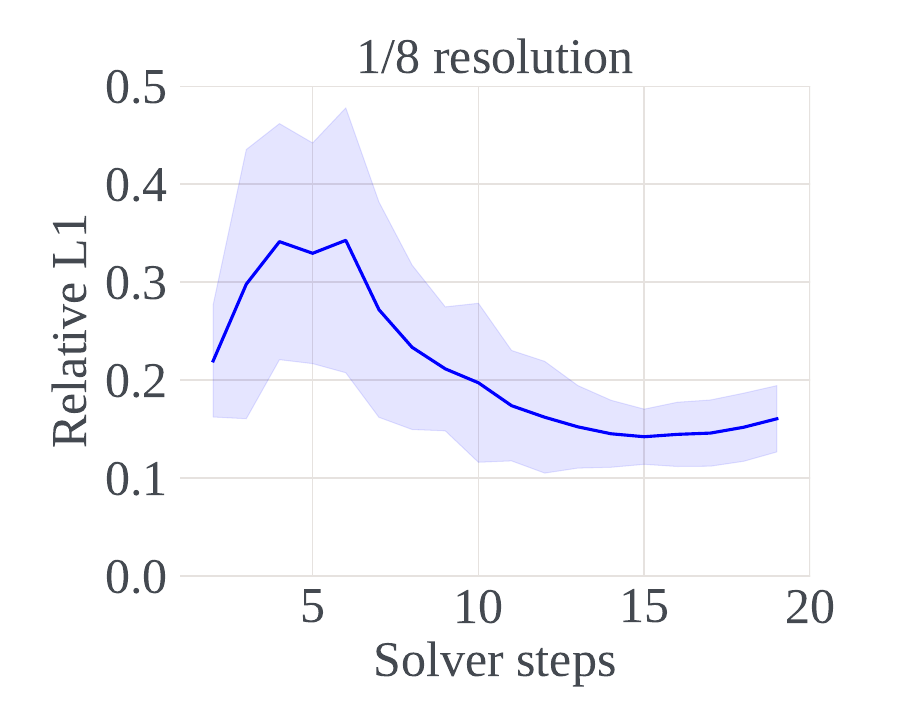} 
    \end{tabular}

\end{subfigure} \\
\footnotesize (a) Layer development during denoising. & 
\footnotesize (b) Per step change for all spatial transformer blocks ($\operatorname{L1}_\text{rel}$).
\end{tabular}
\vspace{-.2cm}
\caption{
\textbf{Overview.} 
We observe, that in diffusion models, not only the intermediate results $x$, but also the internal feature maps change smoothly over time. (a) We visualize output feature maps of two layer blocks within the denoising network via PCA. Structures change smoothly at different rates.
(b) We also observe this smooth layer-wise change when plotting the change in output from one step to the next, averaging over many different prompts and randomly initialized noise. Besides the average, we also show the standard deviation as shaded area. The patterns always remain the same.
\textit{Configuration}: LDM-512, DPM, 20 Steps. 
}
\vspace{-.3cm}
\label{fig:analysis}
\end{figure*}

%% file: sec/3_method.tex
\section{Method}
\label{sec:method}

In this work, we investigate the behavior of the different layers in the diffusion U-Net to develop novel ways of speeding up the image generation process. 
The main insight of our method is that large latent diffusion models contain redundant computations that can be recycled between steps without compromising image quality. 
The key to our approach is to cache the outputs of U-Net blocks to be reused in the remaining diffusion steps.

\subsection{Preliminaries}

In the diffusion model framework, we start from an input image $x_0 \in [-1, 1]^{3 \times H \times W}$. 
For a number of timesteps $t \in [1, T]$, we repeatedly add Gaussian noise $\epsilon_t \sim \mathcal{N}$ 
to the image, to gradually transform it into fully random noise.
\begin{equation}
    x_t = x_{t-1} + \epsilon_t
\end{equation}
\begin{equation}
    x_T \sim \mathcal{N}(0, 1)
\end{equation}

To synthesize novel images, we train a neural network $\Psi(x_t, t)$ to gradually \textit{denoise} a random sample $x_T$.
The neural network can be parameterized in different ways to predict $x_0$, $\epsilon_t$ or $\nabla \log(x_t)~$\cite{song2020score}.
A solver $\Phi$ determines how to exactly compute $x_{t-1}$ from the output of $\Psi$ and $t$.
\begin{equation}
    x_{t-1} = \Phi\left(x_t, t, \Psi\left(x_t, t\right)\right)
\end{equation}

The higher the number of steps is, the higher the visual quality of the image generally becomes.
Determining the number of steps presents users with a trade-off between image quality and speed.

\subsection{Analysis}
\label{ssec:analysis}
\input{figures/emu_qualitative_results}

One of the key limitations of diffusion models is their slow inference speed. 
Existing works often propose new solvers or to distill existing models, so that fewer steps are required to produce high-quality images.
However, both of these directions treat the given neural network as a black box.

In this paper, we move away from the ``black box perspective'' and investigate the \textit{internal} behavior of the neural network $\Psi$ to understand it at a per-layer basis.
This is particularly interesting when considering the temporal component.
To generate an image, we have to perform multiple forward passes, where the input to the network changes only gradually over time.

The neural network $\Psi$ generally consists of multiple blocks of layers $B_i(x_i, s_i)$, $i \in [0, N-1]$, where $N$ is the number of all blocks of the network, $x$ is the output of an earlier block and $s$ is the optional data from a skip connection.
The common U-Net architecture~\cite{ronneberger2015u}, as used in many current works~\cite{rombach2022high, podell2023sdxl, dai2023emu}, is made up of \verb|ResBlock|s, \verb|SpatialTransformer| blocks, and up/downsampling blocks.
\verb|ResBlock|s mostly perform cheap convolutions, while \verb|SpatialTransformer| blocks perform self- and cross-attention operations and are much more costly.

\input{figures/cache_manager_ldm}

A common design theme of such blocks is that they rely on \textit{residual connections}.
Instead of simply passing the results of the layer computations to the next block, the result is combined with the original input of the current block via summation.
This is beneficial, as it allows information (and gradients) to flow more freely through the network~\cite{hardt2016identity}.
Rather than replacing the information, a block \textit{changes} the information that it receives as input.
\begin{align}
    B_i(x, s) &= C_i(x, s) + \operatorname{concat}(x, s)\\
    C_i(x, s) &= \operatorname{layers}_i(\operatorname{concat}(x, s))
\end{align}

To better understand the inner workings of the neural network, we visualize how much the changes the block applies to the input vary over time.
Concretely, we consider two metrics: Relative absolute change $\operatorname{L1}_{\text{rel}}$.
\begin{equation}
    \operatorname{L1}_{\text{rel}}(i, t) = \frac{||C_i(x_t, s_t) - C_i(x_{t-1}, s_{t-1})||_1}{||C_i(x_t, s_t)||_1}
\end{equation}

To get representative results, we generate 32 images from different prompts with 2 random seeds each and report the averaged results in \cref{fig:analysis}.
Further, we visualize selected feature maps. 
We make three \textbf{key observations}:

\textbf{1) Smooth change over time.}
Similarly to the intermediate images during denoising, the blocks change smoothly and gradually over time.
This suggests that there is a clear temporal relation between the outputs of a block.

\textbf{2) Distinct patterns of change.}
The different blocks do not behave uniformly over time.
Rather, they apply a lot of change in certain periods of the denoising process, while they remain inactive in others.
The standard deviation shows that this behavior is consistent over different images and random seeds.
Note that some blocks, for example the blocks at higher resolutions (either very early or very late in the network) change most in the last 20\%, while deeper blocks at lower resolutions change more in the beginning. 

\textbf{3) Small step-to-step difference.}
Almost every block has significant periods during the denoising process, in which its output only changes very little.

\subsection{Block Caching}

We hypothesize that a lot of blocks are performing redundant computations during steps where their outputs change very little.
To reduce the amount of redundant computations and to speed up inference, we propose \textbf{Block Caching}.

Instead of computing new outputs at every step, we reuse the cached outputs from a previous step.
Due to the nature of residual connections, we can perform caching at a per-block level without interfering with the flow of information through the network otherwise.
We can apply our caching technique to almost all recent diffusion model architectures. 

One of the major benefits of \textbf{Block Caching} compared to approaches that reduce the number of steps is that we have a more finegrained control over where we save computation.
While we perform fewer redundant computations, we do not reduce the number of steps that require a lot of precision (\ie where the change is high).
\input{figures/scale_shift_optimization}
\paragraph{Automatic cache schedule.}

Not every block should be cached all the time.
To make a more informed decision about when and where to cache, we rely on the metric described in \cref{ssec:analysis}.
We first evaluate these metrics over a number of random prompts and seeds.
Our intuition is that for any layer block $i$, we retain a cached value, which was computed at time step $t_a$, as long as the accumulated change does not exceed a certain threshold $\delta$.
Once the threshold is exceeded at time step $t_b$, we recompute the block's output.
\begin{equation}
    \sum_{t = t_a}^{t_b - 1} \operatorname{L1}_{\operatorname{rel}}(i, t) \leq \delta < \sum_{t = t_a}^{t_b} \operatorname{L1}_{\operatorname{rel}}(i, t)
\end{equation}
With a lower threshold, the cached values will be refreshed more often, whereas a higher threshold will lead to faster image generation but will affect the appearance of the image more. The threshold $\delta$ can be picked such that it increases inference speed without negatively affecting image quality.

\input{figures/ldm_qualitative_results}
\subsection{Scale-Shift Adjustment}
While caching already works surprisingly well on its own, as shown in \cref{sec:main_results}, we observe that aggressive caching can introduce artifacts into the final image.
We hypothesize that this is due to a misalignment between the cached feature map and the ``original'' feature map at a given timestep.
To enable the model to adjust to using cached values, we introduce a very lightweight \textbf{scale-shift adjustment mechanism} wherever we apply caching.
To this end, we add a timestep-dependent scalar shift and scale parameter for each layer that receives a cached input. 
Concretely, we consider every channel separately, \ie for a feature map of shape $(N \times C \times H \times W)$, we predict a vector of shape $(N \times C)$ for both scale and shift.
This corresponds to a simple linear layer that receives the timestep embedding as input.

We optimize scale and shift on the training set while keeping all other parameters frozen. 
However, optimization of these additional parameters is not trivial.
As we require valid cached values, we cannot directly add noise to an image and train the network to denoise to the original image.

Therefore, we rely on an approach, shown in \cref{fig:scale_shift_optimization}, that is inspired by distillation techniques.
Our model with caching enabled acts as the student, while the same model with caching disabled acts as the teacher.
We first unroll the consecutive steps of the denoising process for the student configuration and generate an image from complete noise.
Then, we perform a second forward pass at every timestep with the teacher configuration, which acts as the training target.
Note that for the teacher, we use the intermediate steps from the student's trajectory as input rather than unrolling the teacher. 
Otherwise, the teacher might take a different trajectory (leading to a different final output), which then is not useful as a training target.

This optimization is very resource-friendly, as the teacher and student can use the same weights, saving GPU memory, and we only optimize a small number of extra parameters, while keeping the parameters of the original model the same.
During inference, the multiplication and addition with scale and shift parameters have no noticeable effect on the inference speed but improve image quality as shown in \cref{sec:main_results}.

%% file: figures/emu_qualitative_results.tex
\begin{figure*}
\centering
\includegraphics[trim={0cm .2cm 0cm 0cm},clip,width=\linewidth]{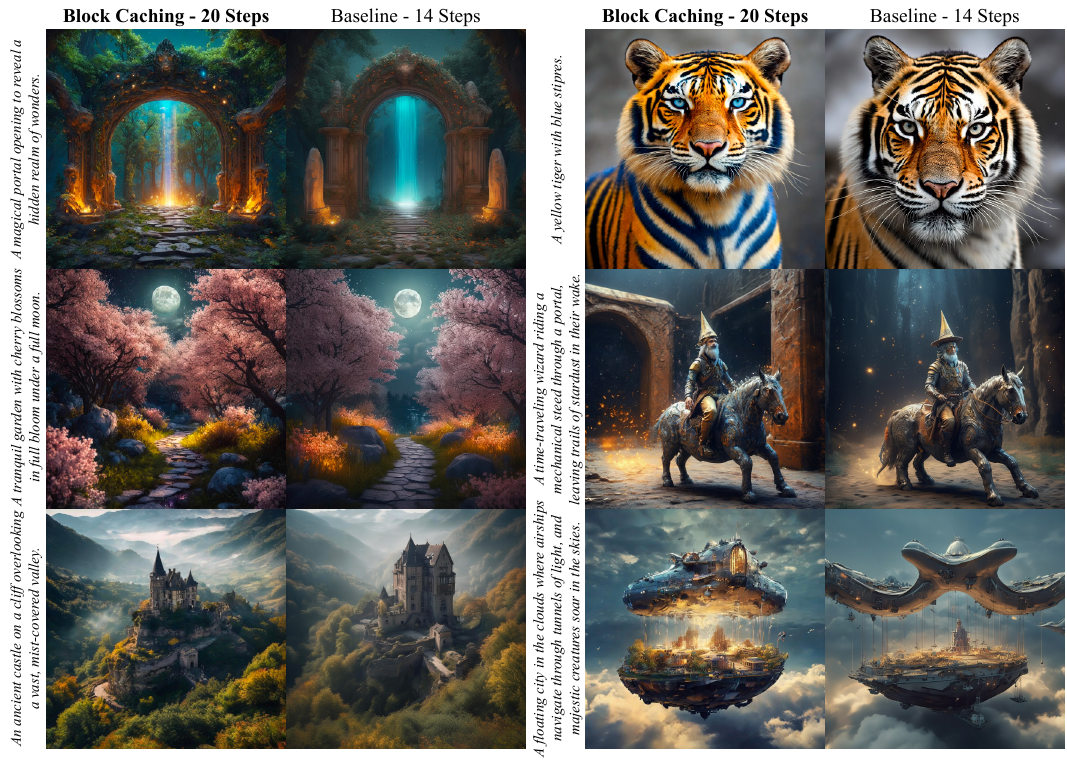}
\vspace{-.6cm}
\caption{
\textbf{Qualitative Results for EMU-768.} 
With identical inference speed, our caching technique produces finer details and more vibrant colors. For more results refer to the supplementary material. \textit{Configuration}: DPM, Block caching with 20 steps vs Baseline with 14 steps.
}
\vspace{-.3cm}
\label{fig:emu_qualitative_results}
\end{figure*}

%% file: figures/cache_manager_ldm.tex
\begin{figure}[t]
\centering
\includegraphics[trim={1.0cm 0cm 0.0cm 1cm},clip,width=\linewidth]%
{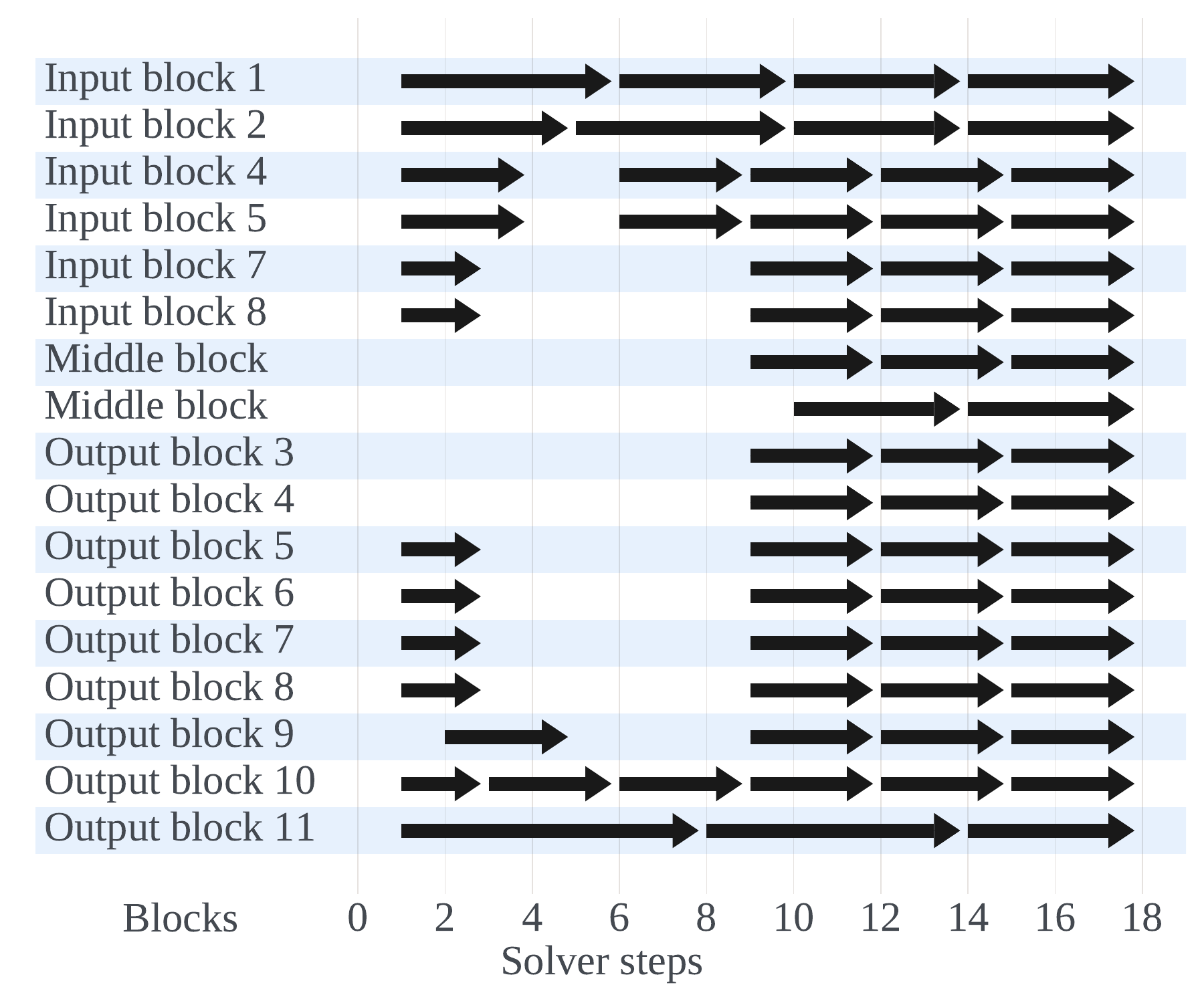}
\vspace{-.7cm}
\caption{
\textbf{Caching Schedule for LDM-512 at 20 steps with DPM.} Each arrow represents the cache lifetime of a spatial transformer block. For the duration of an arrow, the spatial transformer block reuses the cached result computed at the beginning of the arrow. E.g., \textit{Input block 1} only computes the result at step 1, 6, 10, 14 and 18 and uses the cached value otherwise. 
}

\label{fig:caching_schedule}
\vspace{-.2cm}
\end{figure}

%% file: figures/scale_shift_optimization.tex
\begin{figure}[t]
\centering
\includegraphics[trim={ 0cm 0cm 0.0cm 0cm},width=\columnwidth]{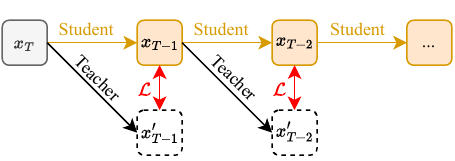}
\caption{
\textbf{Scale Shift Optimization.} The student network copies and freezes the weights of the teacher and has additional scale and shift parameters per block. These parameters are optimized to match the teacher output per block and step.
}
\label{fig:scale_shift_optimization}
\end{figure}

%% file: figures/ldm_qualitative_results.tex
\begin{figure*}
\centering
\includegraphics[trim={0cm 0cm 0cm 0cm},clip,width=.90\linewidth]{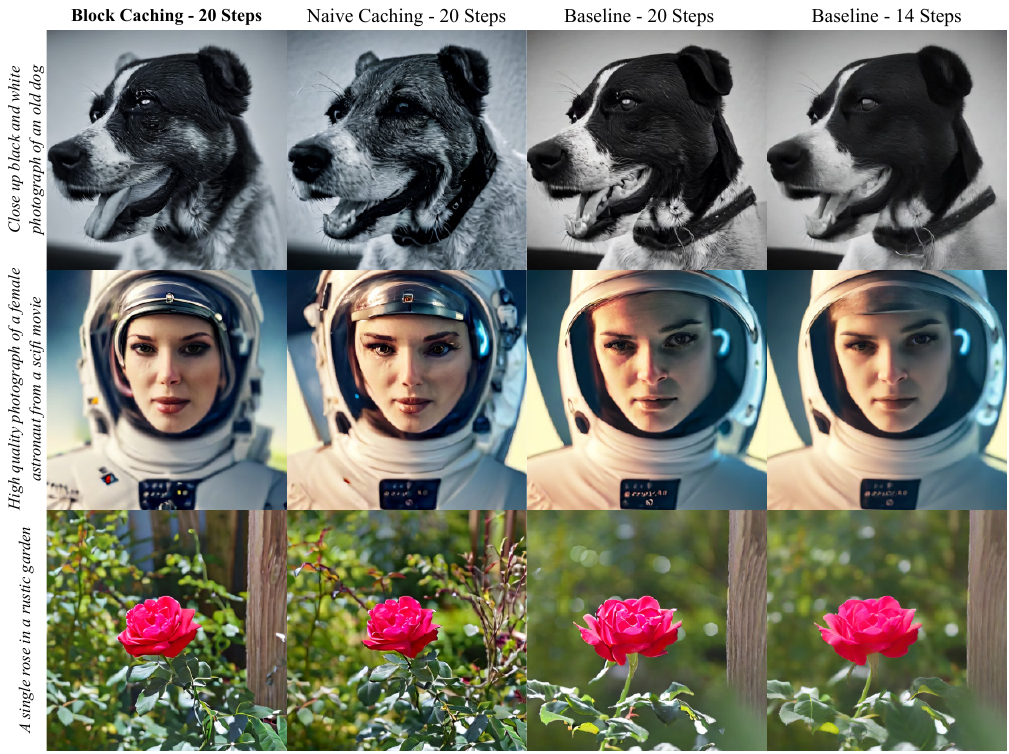}
\caption{
\textbf{Qualitative Results for LDM-512.} Our method often provides richer colors and finer details. Through our scale-shift adjustment, we avoid artifacts that are visible when naively applying block caching. More qualitative results for DPM and DDIM can be found in the supplementary material. \textit{Configuration}: DPM, Block caching with 20 steps vs Baseline with 14 steps.
}
\label{fig:ldm_qualitative_results}
\end{figure*}

%% file: sec/4_experiments.tex
\section{Experiments}

In the following, we first demonstrate the general potential of our \textbf{Block Caching} technique and then analyze it in more detail through several ablation studies.

\subsection{Experimental Setup}

Our proposed method is general and can be applied to most recent diffusion models.
In order to give a good overview, we conduct our experiments mainly on two models that represent light and heavy computational demands:
\begin{itemize}
    \item \textbf{LDM-512} \cite{rombach2022high}, a popular diffusion model with 900M parameters, that generates images at a $512 \times 512$ resolution, retrained on internal Shutterstock images.
    \item \textbf{EMU-768} \cite{dai2023emu}, a state-of-the-art model with 2.7B parameters, which can produce photorealistic images at a resolution of $768\times 768$.
\end{itemize}

For both models, we use classifier-free guidance \cite{ho2021classifier} with a guidance strength of $5.0$ and do not use any other performance-enhancing techniques.
We run inference in $\operatorname{bfloat16}$ type and measure the latency on a single Nvidia A100 GPU.
For the optimization of the scale-shift adjustment parameters, we perform 15k training iterations on eight A100 GPUs.
Depending on the model and the number of denoising steps, this takes between 12 and 48 hours.

\subsection{Accelerating Inference through Caching}
\label{sec:main_results}
Our proposed caching technique can be viewed from two perspectives:
1) Given a fixed number of steps, caching allows us to accelerate the image generation process without decreasing quality.
2) Given a fixed computational budget, we can perform more steps when using caching, and therefore obtain better image quality than performing fewer steps without caching.

To demonstrate the flexibility of our approach, we consider two common inference settings:
(i) Many approaches perform 50 denoising steps by default.
Therefore, we apply caching with 50 solver steps and achieve the same latency as the 30 steps of the baseline model.
(ii) By using modern solvers like DPM \cite{lu2022dpm} or DDIM \cite{song2020denoising}, it is possible to generate realistic-looking images with as few as 20 steps.
If we apply caching with 20 solver steps, we can reduce the inference latency to an equivalent of performing 14 steps with the non-cached baseline model.

\paragraph{Analysis of LDM-512.}
\input{tables/ldm_quantitative_results}
\input{tables/emu_human_eval}

We begin by performing a thorough qualitative and quantitative analysis of the LDM-512 model.
After computing the layer block statistics for the automatic cache configuration, we find that a change threshold of $\delta = 0.5$ gives us the desired speedup.
The resulting caching schedule is visualized in \cref{fig:caching_schedule}.
As can be observed in the plots with relative feature changes (\cref{fig:analysis}), we can aggressively cache the early and late blocks.
On the other hand, the activations of the deeper blocks change faster, especially in the first half of the denoising process, and should therefore only be cached conservatively.

The results in \cref{tab:ldm_latency_fid} demonstrate that for both DPM and DDIM, the proposed caching with 20 steps significantly improves the FID value compared to the 14-step baseline, while being slightly faster. Similarly, 50 steps with caching outperforms the 30-step baseline, while maintaining a comparable latency.
Moreover, our scale-shift adjustment mechanism further enhances the results. Notably, this full configuration even outperforms the 20-step and 50-step baselines. 
We hypothesize that caching introduces a slight momentum in the denoising trajectory due to the delayed updates in cached values, resulting in more pronounced features in the final output image.

Qualitative results can be seen in \cref{fig:ldm_qualitative_results}.
Our full model (caching + scale-shift adjustment) produces more crisp and vibrant images with significantly more details when compared to the 14-step baseline.
This can be explained by the fact that when performing only 14 steps, the model makes steps that are too big to add meaningful details to the image.
Caching without scale-shift adjustment also yields images with more detail compared to the baseline.
However, we often observe local artifacts, which are particularly noticeable in the image backgrounds.
These artifacts appear like overly-emphasized style features.
The application of our scale-shift adjustment effectively mitigates these effects.
\paragraph{Analysis of EMU-768.}

To demonstrate the generality of our proposed approach, we also apply caching and scale-shift adjustment to the EMU-768 model under the same settings as for LDM-512.
As can be seen in \cref{fig:emu_qualitative_results}, we achieve a very similar effect: 
The generated images are much more detailed and more vibrant, compared to the baseline.
This is also confirmed by a human eval study, in which we asked 12 independent annotators to compare the visual appeal of images generated for the prompts from Open User Input (OUI) Prompts \cite{dai2023emu} and PartiPrompts\cite{yu2022scaling} for different configurations.
Specifically, we compared different configurations with the same latency for different samplers and collected 1320 votes in total.
As reported in \cref{tab:emu_human_eval}, our proposed caching technique is clearly preferred over the baseline in every run.
Note that for many prompts, both images have very high quality, leading to a high rate in ties.
This study shows that caching can be applied to a wide range of different models, samplers and step counts.

\paragraph{Effects of more aggressive caching.}
\input{figures/cache_threshold_sweep}
The extent to which the model caches results is controlled by the parameter $\delta$.  
The higher $\delta$, the longer the cache lifetime and the less frequent block outputs are recomputed. \cref{fig:cache_threshold} shows synthesized images for varying $\delta$ values along with the corresponding inference speed. 
Although a higher $\delta$ leads to faster inference, the quality of the final image deteriorates when block outputs are recomputed too infrequently.
We find that $\delta=0.5$ not only provides a significant speedup by 1.5$\times$ but also improves the image quality, thereby achieving the optimal trade-off (see Tab. \ref{tab:ldm_latency_fid}).

\paragraph{Difficulty of Caching ResBlocks.}
\input{figures/resblock_caching}

As described above, we only cache \verb|SpatialTransformer| blocks and not \verb|ResBlock|s.
This design choice is grounded in the observation, that  \verb|ResBlocks| change much less smoothly compared to  \verb|SpatialTransformer| blocks.
In particular, \verb|ResBlocks| are very important for generating local details in the image.
To test this, we generate images where we only cache \verb|ResBlocks| and leave \verb|SpatialTransformer| blocks untouched.
As can be seen in \cref{fig:resblock_caching}, even to gain a speedup of as low as 5\%, the image quality deteriorates significantly.

%% file: tables/ldm_quantitative_results.tex
\setlength{\tabcolsep}{0.3em} 
\renewcommand{\arraystretch}{1.0} 

\begin{table}[]

\centering
\small
\begin{tabular}{c|c|cc|ccc} 
\toprule
Solver & Steps & Caching & SS & FID $\downarrow$ & Img/s $\uparrow$&  Speedup $\uparrow$\hspace{-.2cm} \\ 
\midrule \midrule

\multirow{4}{*}{ \begin{tabular}{c} DPM \\ \cite{lu2022dpm}\end{tabular}} & 20  & & & \underline{17.15} & 2.17 & 1.00$\times$ \\
        & 14  & & & 18.67 & 3.10 & 1.43$\times$  \\
        & 20 & \checkmark   & & 17.58 & \textbf{3.64} & \textbf{1.68$\times$} \\
        & \textbf{20} & \checkmark & \checkmark & \textbf{15.95} & \underline{3.59}  & \underline{1.65$\times$}  \\
\midrule
\multirow{4}{*}{\begin{tabular}{c} DDIM \\ \cite{song2020denoising} \end{tabular}}  & 20 &  &  & 17.43  & 2.17 & 1.00$\times$  \\
        & 14  & &  & 17.11 & 3.10  & 1.43$\times$ \\
        & 20 & \checkmark  &  & \underline{16.52} & \textbf{3.48} & \textbf{1.60$\times$} \\
        & \textbf{20} & \checkmark & \checkmark & \textbf{16.02} & \underline{3.45} & \underline{1.58$\times$} \\

\midrule \midrule

\multirow{4}{*}{ \begin{tabular}{c} DPM \\ \cite{lu2022dpm}\end{tabular}} & 50  & & & 17.44 & 0.87 & 1.00$\times$ \\
        & 30  & & & \underline{17.21} & 1.46 & 1.67$\times$  \\
        & 50 & \checkmark   & & 17.23 & \textbf{1.61} & \textbf{1.85$\times$} \\
        & \textbf{50} & \checkmark & \checkmark & \textbf{15.18} & \underline{1.59}  & \underline{1.82$\times$}  \\
\midrule
\multirow{4}{*}{\begin{tabular}{c} DDIM \\ \cite{song2020denoising} \end{tabular}}  & 50 &  &  & 17.76  & 0.87 & 1.00$\times$  \\
        & 30  & &  & 17.42 & 1.46  & 1.67$\times$ \\
        & 50 & \checkmark  &  & \underline{16.65} & \textbf{1.59} & \textbf{1.82$\times$} \\
        & \textbf{50} & \checkmark & \checkmark & \textbf{15.15} & \underline{1.56} & \underline{1.79$\times$} \\

\bottomrule
\end{tabular}
\caption{\textbf{LDM-512 FID and Throughput Measurements.} For different solvers, we test our caching technique against baselines with 1) the same number of steps or 2) the same latency. In all cases, our proposed approach achieves significant speedup while improving visual quality as measured by FID on a COCO subset removing all faces (for privacy reasons).
\textbf{Legend}: SS = Scale-shift adjustment, Img/s.\ = Images per second.}
\label{tab:ldm_latency_fid}
\vspace{-.5cm}
\end{table}

%% file: tables/emu_human_eval.tex
\setlength{\tabcolsep}{0.5em}
\renewcommand{\arraystretch}{1.2}

\begin{table}[t] 
\centering

\small
\begin{tabular}{c|cc|ccc}
\toprule
\multirow{2}{*}{Solver} & \multicolumn{2}{c|}{Steps $_\textit{(Img/s)}$} & \multicolumn{3}{c}{Votes (in \%)} \\
 & Caching & Baseline  & Win & Tie & Lose \\
\midrule\midrule
DPM & 20 $_{(0.28)}$ & 14 $_{(0.27)}$ & 34.7 & 36.9 & 28.4 \\
DDIM & 20 $_{(0.28)}$ & 14 $_{(0.27)}$ & 28.0 & 48.8 & 23.2 \\
\midrule
DPM & 50 $_{(0.14)}$ & 30 $_{(0.13)}$& 27.8 & 54.3 & 17.9 \\
DDIM & 50 $_{(0.13)}$ & 30 $_{(0.13)}$ & 29.7 & 46.8 & 23.5 \\
\bottomrule
\end{tabular} 
\caption{\textbf{EMU-768 Visual Appeal Human Evaluation.}
We present the percentages of votes indicating a win, tie, or loss for our method in comparison to the baseline. This is evaluated across various solvers and number of steps. In every comparison, both the caching and baseline configuration have roughly the same inference speed (reported as images per second).
}
\label{tab:emu_human_eval}
\vspace{-.3cm}
\end{table}

%% file: figures/cache_threshold_sweep.tex
\begin{figure}
\vspace{-0.7em}
\centering
\begin{tabular}{l r}
\begin{subfigure}[t]{0.52\linewidth}
\includegraphics[trim={0cm 0cm 0cm 0cm},clip,width=\linewidth]{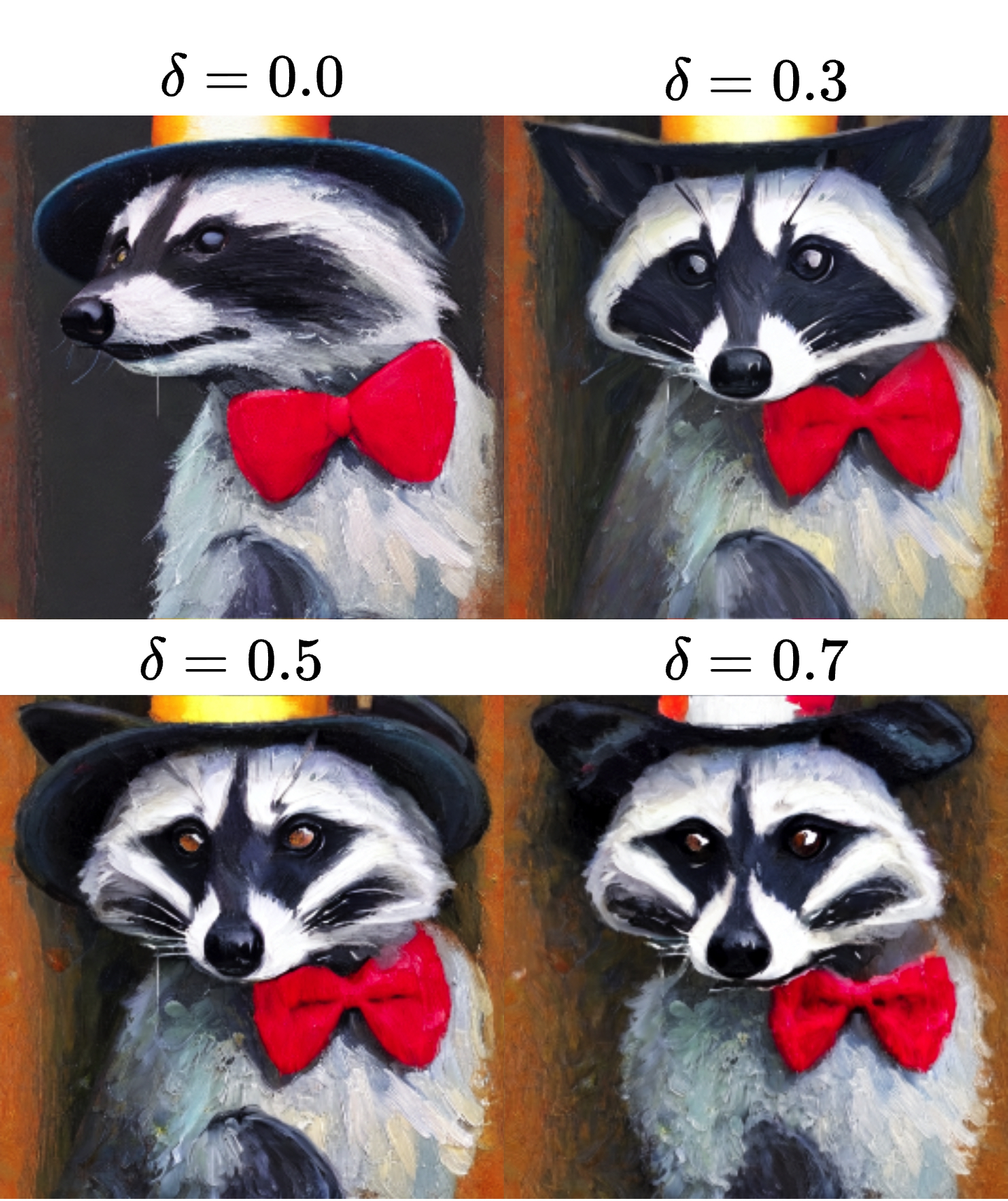} 
\end{subfigure}
& 
\begin{subfigure}[b]{0.46\linewidth}

\includegraphics[trim={0cm 1.5cm 0.0cm 1cm},clip,width=\linewidth]{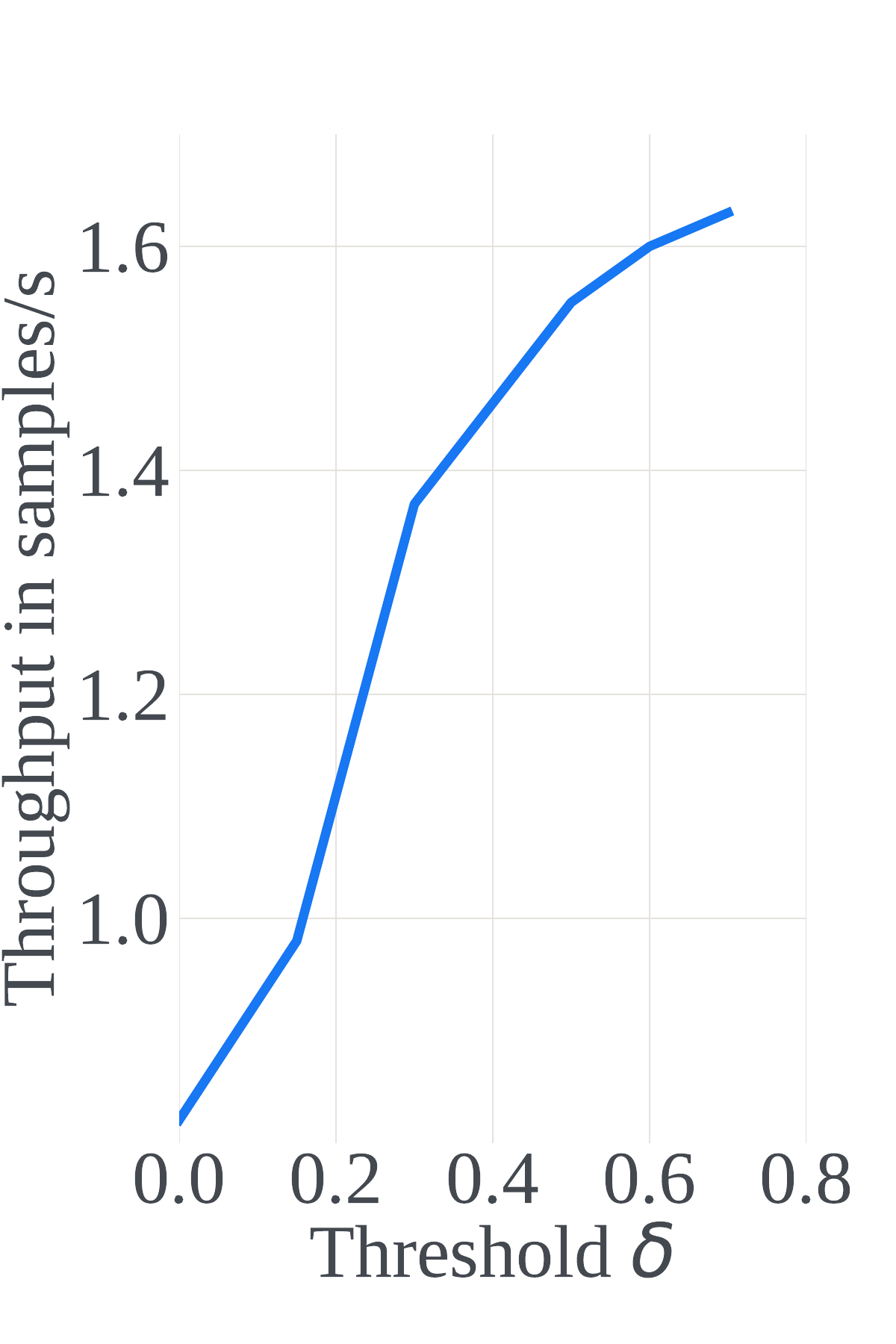}
\end{subfigure}
\end{tabular}
\caption{
\textbf{Effect of Cache Threshold $\delta$.} Left: Generated image for different $\delta$. Right: Inference speed vs. $\delta$. The higher $\delta$, the more blocks are cached, resulting in faster inference. $\delta=0.5$ gives a 1.5x speedup and the best visual quality. \textit{Configuration}: DPM, LDM-512, Block caching with 50 steps. 
}
\label{fig:cache_threshold}
\vspace{-.3cm}
\end{figure}

%% file: figures/resblock_caching.tex
\begin{figure}[t]
\centering
\includegraphics[trim={0cm 0cm 0cm 0cm},clip,width=0.95\linewidth]{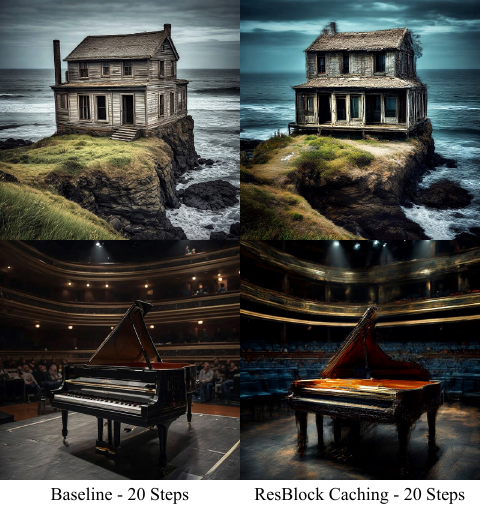}
\caption{
\textbf{Effect of Caching ResBlocks.} 
Caching ResBlocks instead of spatial transformer blocks results in fewer details and inferior image quality, while achieving only a small speedup of 5\%. \textit{Configuration}: DPM, EMU-768, Block caching with 20 steps.
}
\label{fig:resblock_caching}
\vspace{-.3cm}
\end{figure}

%% file: sec/5_conclusion.tex
\section{Conclusion}

In this paper, we first analyzed the inner workings of the denoising network, moving away from the common perspective of considering diffusion models as black boxes.
Leveraging the insights from our analysis, we proposed the \textbf{Block Caching} technique. It reduces the redundant computations during inference of the diffusion models and significantly speeds up the image generation process by a factor of \textbf{1.5$\times$}-\textbf{1.8$\times$} at a minimal loss of image quality.
To showcase the adaptability of our approach, we performed experiments on LDM and EMU models with a parameter range from 900M to 2.7B. 
We tested our approach in different inference settings by varying solvers and number of steps. 
Our technique generates more vibrant images with more fine-grained details when compared to naively reducing the number of solver steps for the baseline model to match the compute budget. 
We confirmed our findings quantitatively by computing the FID and by human evaluation.

%% file: sec/X_suppl.tex
\maketitlesupplementary

\renewcommand{\thesection}{\Alph{section}}
\setcounter{section}{0}   
\setcounter{figure}{0}   
\setcounter{table}{0}

\section*{Supplementary Material}
\label{sec:supp}
In this supplementary material, we provide 
\begin{enumerate}

    \item thoughts on future work in \cref{sec:future_work}
    \item an overview of the limitations of our method in \cref{sec:limitations}
    \item thoughts on ethical considerations and safety in \cref{sec:ethics}
    \item additional figures for \textbf{qualitative results}, \textbf{change metric plots}, and \textbf{caching schedules} in \cref{sec:supp_additional_figures}
\end{enumerate}

\section{Future Work}
\label{sec:future_work}

There are several directions for future work.
First, we believe that the use of step-to-step change metrics is not limited to caching, but could / should also benefit e.g. finding a better network architecture or a better noise schedule.
Secondly, we find that the effect of scale-shift adjustment can be quite significant on the overall structure and visual appeal of the image. 
It could be possible to use a similar technique for finetuning with human in the loop to make the model adhere more to the preference of the user without having to change the training data.
Finally, it would be interesting if caching could be integrated into a network architecture even before training. 
This could not only improve the results of the final model, but also speed up training.

\section{Limitations}
\label{sec:limitations}

While our method achieves good results, some noteworthy weaknesses remain.
We observe that while the scale-shift adjustment improves results and reduces artifacts, it sometimes changes the identity of the image more than reducing the number of steps or using naive caching would.
Furthermore, 
finding the perfect threshold for auto configuration can take time, as the model is sensitive to certain changes in the caching schedule.
\
We recommend playing around with small variations of the desired threshold to obtain the perfect schedule.

\section{Ethical Considerations \& Safety}
\label{sec:ethics}

We do not introduce new image data to these model and the optimization scheme for scale-shift adjustment only requires prompts.
Therefore, we believe that our technique does not introduce ethical or legal challenges beyond the model on which we apply our technique.

For safety considerations, it should be noted that scale-shift adjustment, while still following the prompt, can change the identities in the image slightly. 
This aspect might make an additional safety check necessary when deploying models with block caching.

\section{Additional Figures}
\label{sec:supp_additional_figures}

\textbf{Additional Qualitative Results.}
We show additional results for all configurations mentioned in the main paper. For all configurations, we show our caching technique with and without scale-shift adjustment, a slower baseline with the same number of steps, and a baseline with the same latency as ours (by reducing the number of steps).
\newline $ $
\newline
\noindent
\textbf{Additional Change Plots.}
For all above mentioned configurations, we show the step-to-step change per layer block averaged over 32 forward passes and two random seeds each measured via the $\operatorname{L1}_\text{rel}$ metric.
This corresponds to Fig. 2 b) in the main paper.
\newline $ $
\newline
\noindent
\textbf{Additional Caching Schedules,}
Finally, we also show all the caching schedules, which are automatically derived from the change measurements mentioned above. 
\newline $ $
\newline
An overview of the figures is provided by 
\cref{tab:supp_figure_overview}  
\input{tables_supp/figure_overview}

\input{figures_supp/emu_dpm20_qualitative}
\newpage
\input{figures_supp/emu_ddim20_qualitative}
\newpage
\input{figures_supp/emu_dpm50_qualitative}
\newpage
\input{figures_supp/emu_ddim50_qualitative}
\newpage
\input{figures_supp/ldm_dpm20_qualitative}
\newpage
\input{figures_supp/ldm_ddim20_qualitative}
\newpage
\input{figures_supp/ldm_dpm50_qualitative}
\newpage
\input{figures_supp/ldm_ddim50_qualitative}
\newpage
\input{figures_supp/change_metrics}
\newpage
\input{figures_supp/cache_schedules}

%% file: tables_supp/figure_overview.tex
\begin{table}[h]
\small
\centering
\begin{tabular}{cccccc}
\toprule
\textit{Model} & \textit{Steps} & \textit{Solver} & \textit{Quali.} & \textit{Change} & \textit{Schedule}\\
\midrule
\multirow{4}{*}{EMU-768} & \multirow{2}{*}{20 vs 14} & DPM & \cref{fig:supp_emu_dpm20} & \cref{fig:supp_change_emu_dpm20} & \cref{fig:supp_schedule_emu_dpm20}\\
 &  & DDIM & \cref{fig:supp_emu_ddim20} & \cref{fig:supp_change_emu_ddim20} & \cref{fig:supp_schedule_emu_ddim20} \\
 & \multirow{2}{*}{50 vs 30} & DPM & \cref{fig:supp_emu_dpm50} & \cref{fig:supp_change_emu_dpm50} &  \cref{fig:supp_schedule_emu_dpm50}\\
 & & DDIM & \cref{fig:supp_emu_ddim50} & \cref{fig:supp_change_emu_ddim50} & \cref{fig:supp_schedule_emu_ddim50}\\
 \midrule
\multirow{4}{*}{LDM-512} & \multirow{2}{*}{20 vs 14} & DPM & \cref{fig:supp_ldm_dpm20} & \cref{fig:supp_change_ldm_dpm20} & \cref{fig:supp_schedule_ldm_dpm20}\\
 &  & DDIM & \cref{fig:supp_ldm_ddim20} & \cref{fig:supp_change_ldm_ddim20} & \cref{fig:supp_schedule_ldm_ddim20}\\
 & \multirow{2}{*}{50 vs 30} & DPM & \cref{fig:supp_ldm_dpm50} & \cref{fig:supp_change_ldm_dpm50} & \cref{fig:supp_schedule_ldm_dpm50}\\
 & & DDIM & \cref{fig:supp_ldm_ddim50} & \cref{fig:supp_change_ldm_ddim50} & \cref{fig:supp_schedule_ldm_ddim50}\\
 \bottomrule
\end{tabular}
\caption{\textbf{Additional Figures Overview.} \textit{Quali.}: Qualitative results, \textit{Change}: Change metric plots, \textit{Schedule}: Chaching schedule}
\label{tab:supp_figure_overview}

\end{table}

%% file: figures_supp/emu_dpm20_qualitative.tex
\begin{figure*}
\centering
\includegraphics[trim={0cm 0cm 0cm 0cm},clip,width=\linewidth]{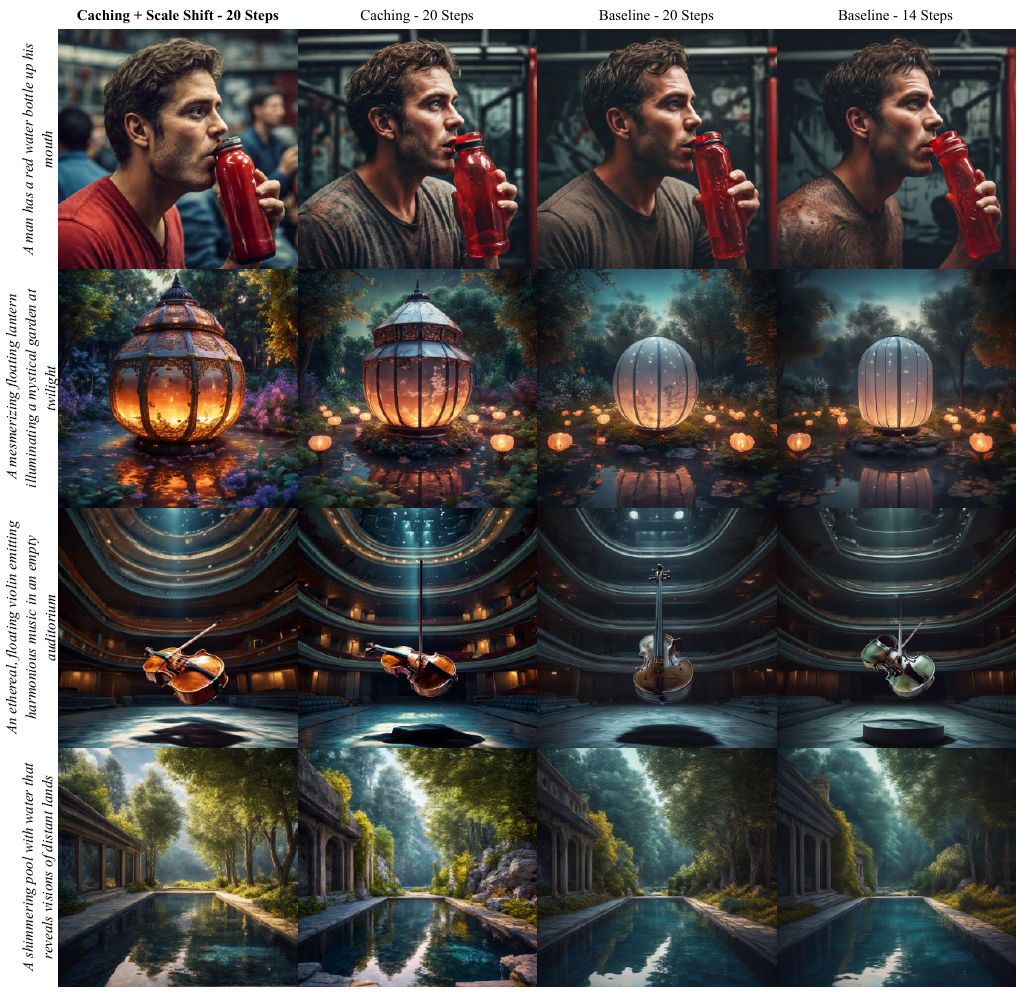}
\caption{
\textbf{Qualitative Results for EMU-768 - DPM 20 Steps.}}
\label{fig:supp_emu_dpm20}
\end{figure*}

%% file: figures_supp/emu_ddim20_qualitative.tex
\begin{figure*}
\centering
\includegraphics[trim={0cm 0cm 0cm 0cm},clip,width=\linewidth]{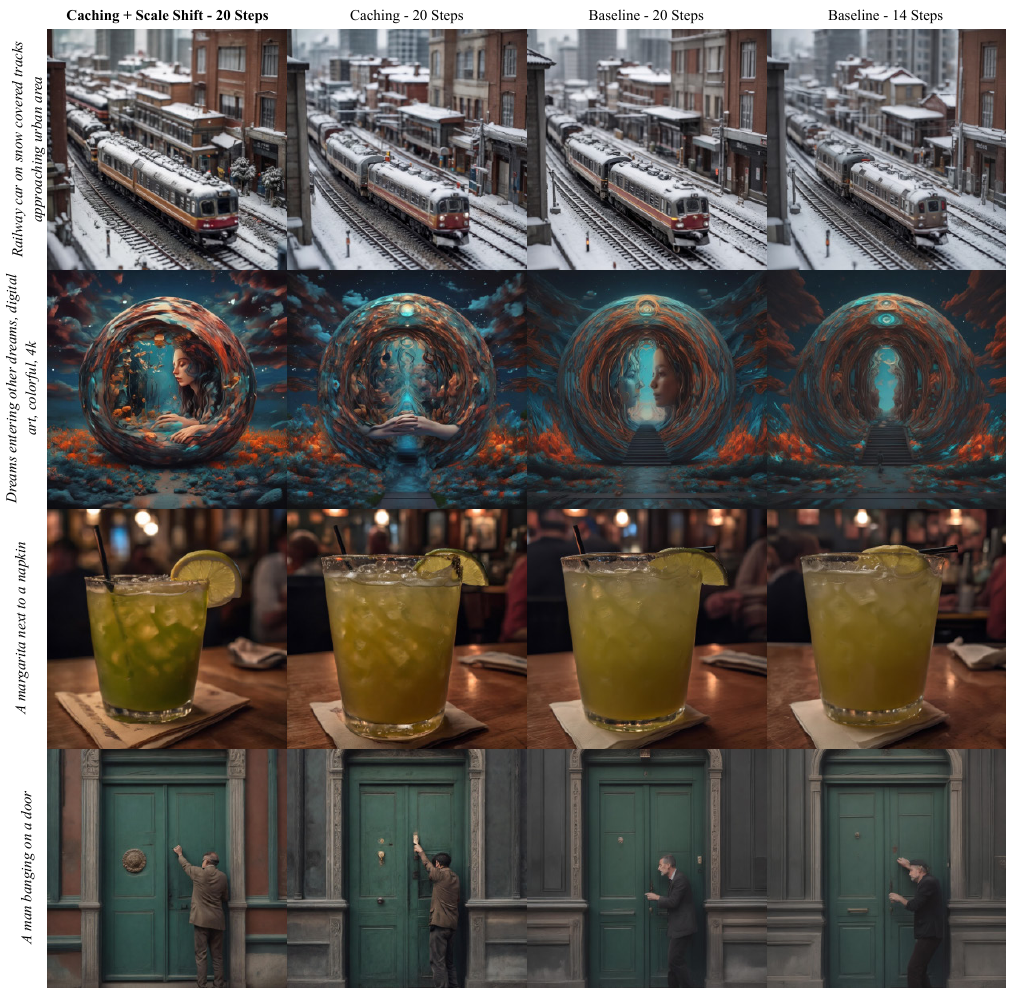}
\caption{
\textbf{Qualitative Results for EMU-768 - DDIM 20 Steps.}}
\label{fig:supp_emu_ddim20}
\end{figure*}

%% file: figures_supp/emu_dpm50_qualitative.tex
\begin{figure*}
\centering
\includegraphics[trim={0cm 0cm 0cm 0cm},clip,width=\linewidth]{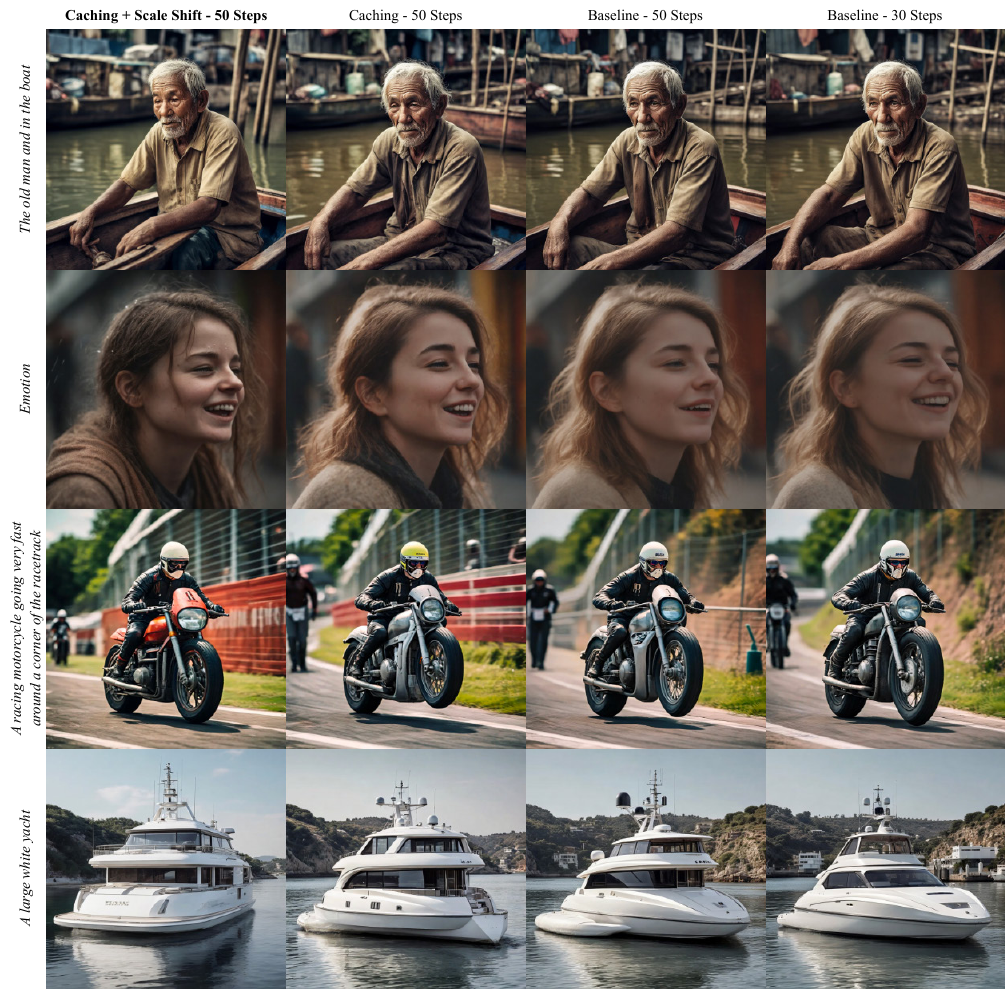}
\caption{
\textbf{Qualitative Results for EMU-768 - DPM 50 Steps.}}
\label{fig:supp_emu_dpm50}
\end{figure*}

%% file: figures_supp/emu_ddim50_qualitative.tex
\begin{figure*}
\centering
\includegraphics[trim={0cm 0cm 0cm 0cm},clip,width=\linewidth]{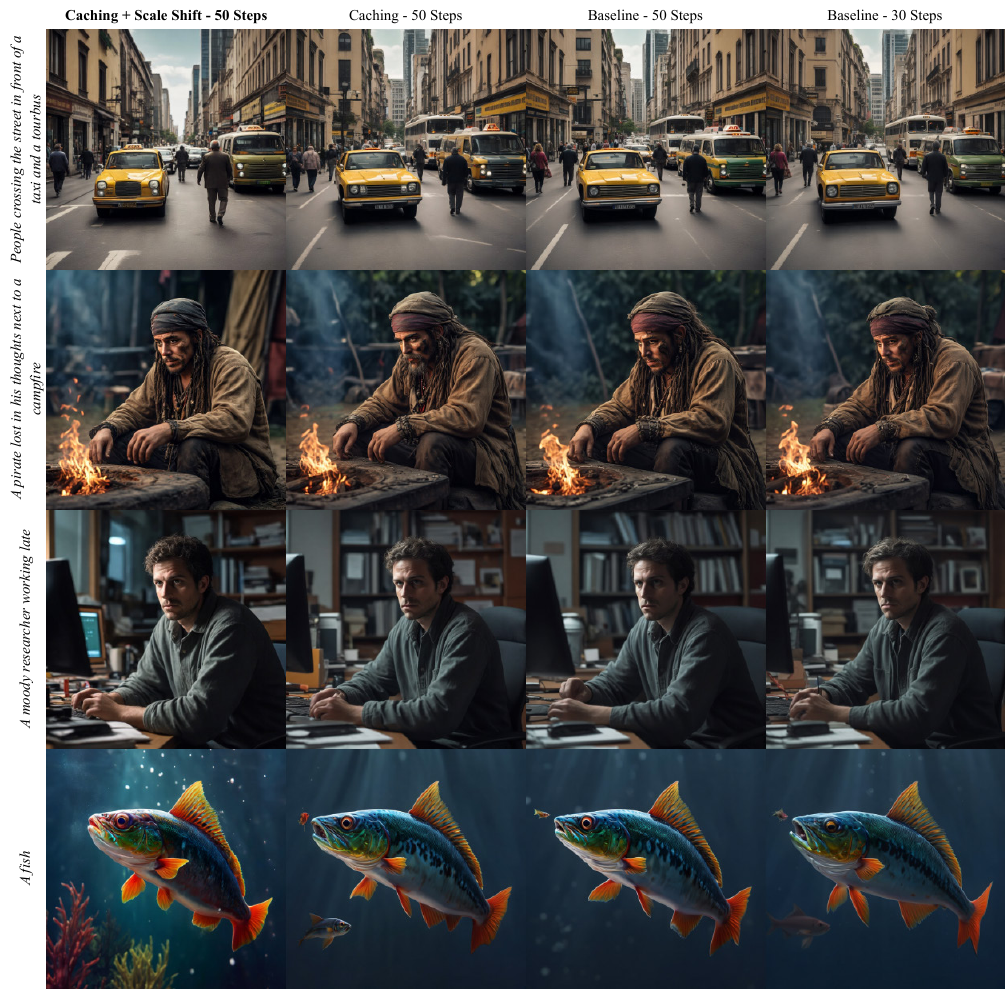}
\caption{
\textbf{Qualitative Results for EMU-768 - DDIM 50 Steps.}}
\label{fig:supp_emu_ddim50}
\end{figure*}

%% file: figures_supp/ldm_dpm20_qualitative.tex
\begin{figure*}
\centering
\includegraphics[trim={0cm 0cm 0cm 0cm},clip,width=\linewidth]{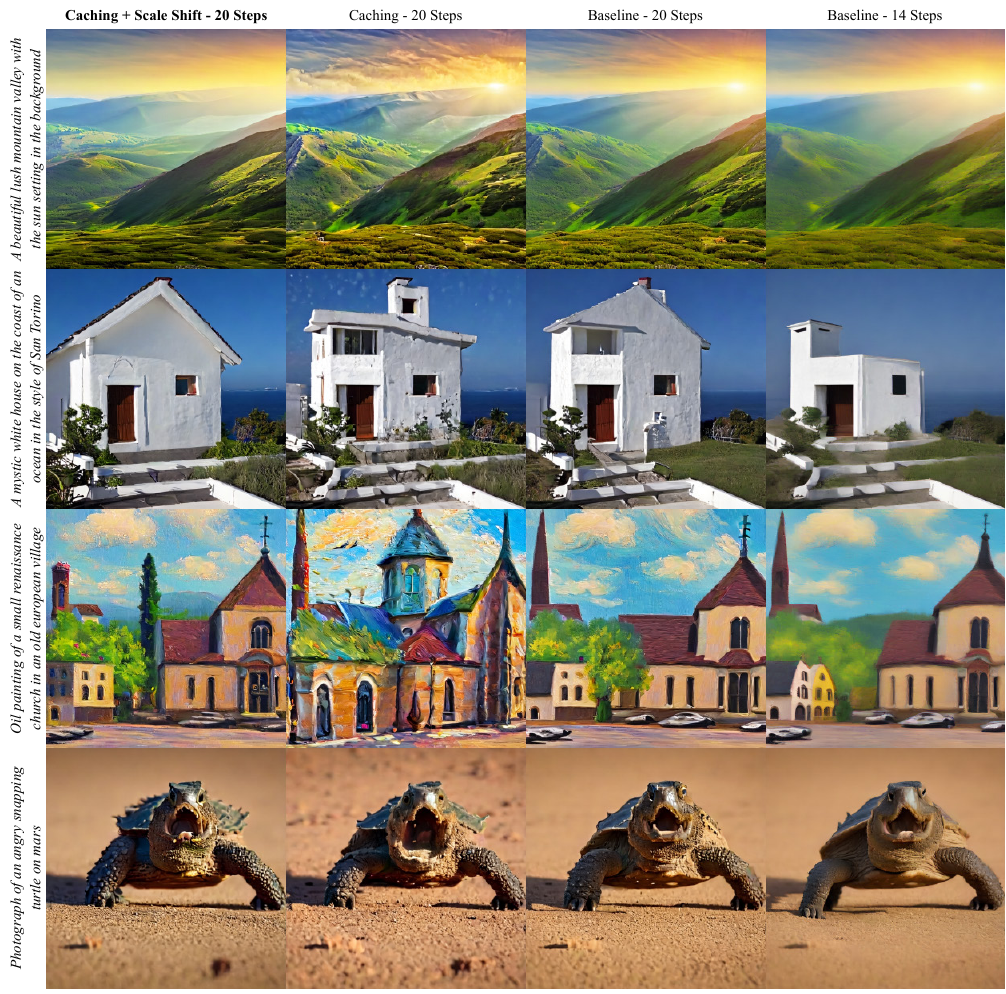}
\caption{
\textbf{Qualitative Results for LDM-512 - DPM 20 Steps.}}
\label{fig:supp_ldm_dpm20}
\end{figure*}

%% file: figures_supp/ldm_ddim20_qualitative.tex
\begin{figure*}
\centering
\includegraphics[trim={0cm 0cm 0cm 0cm},clip,width=\linewidth]{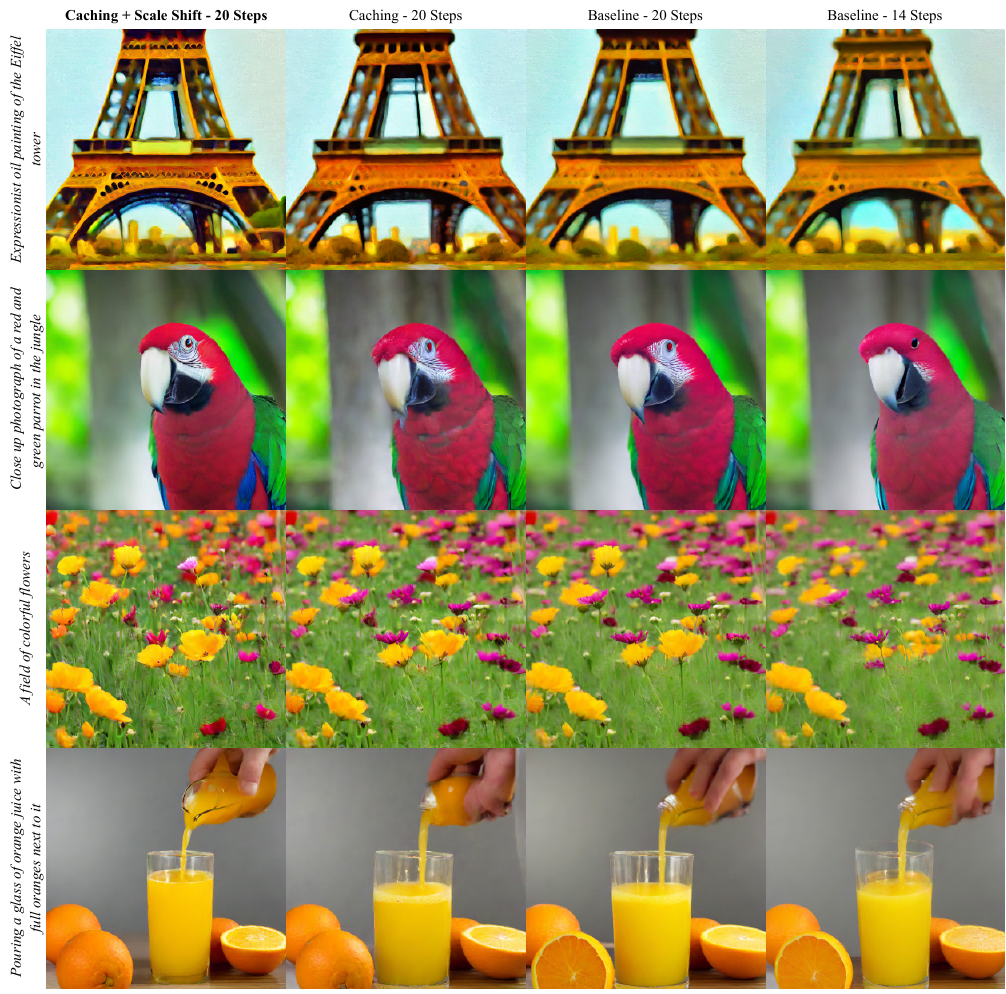}
\caption{
\textbf{Qualitative Results for LDM-512 - DDIM 20 Steps.}}
\label{fig:supp_ldm_ddim20}
\end{figure*}

%% file: figures_supp/ldm_dpm50_qualitative.tex
\begin{figure*}
\centering
\includegraphics[trim={0cm 0cm 0cm 0cm},clip,width=.90\linewidth]{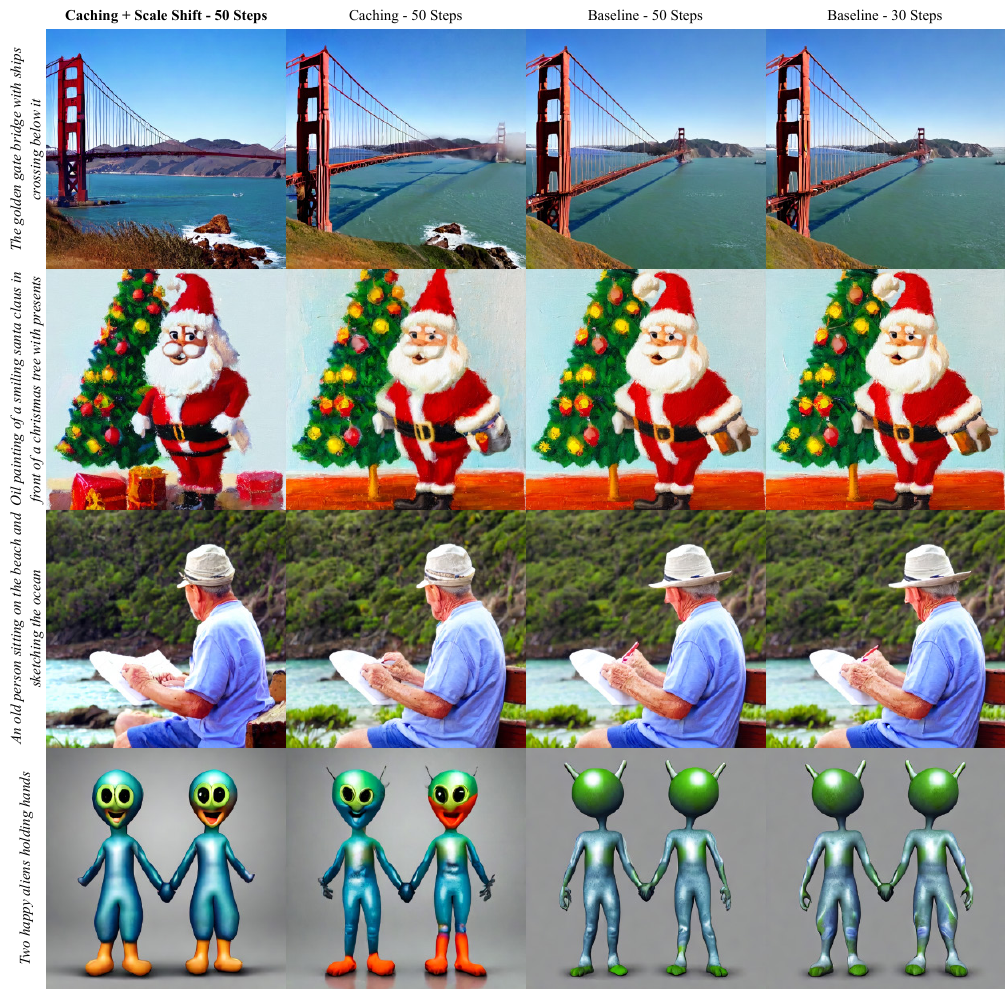}
\caption{
\textbf{Qualitative Results for LDM-512 - DPM 50 Steps.}}
\label{fig:supp_ldm_dpm50}
\end{figure*}

%% file: figures_supp/ldm_ddim50_qualitative.tex
\begin{figure*}
\centering
\includegraphics[trim={0cm 0cm 0cm 0cm},clip,width=\linewidth]{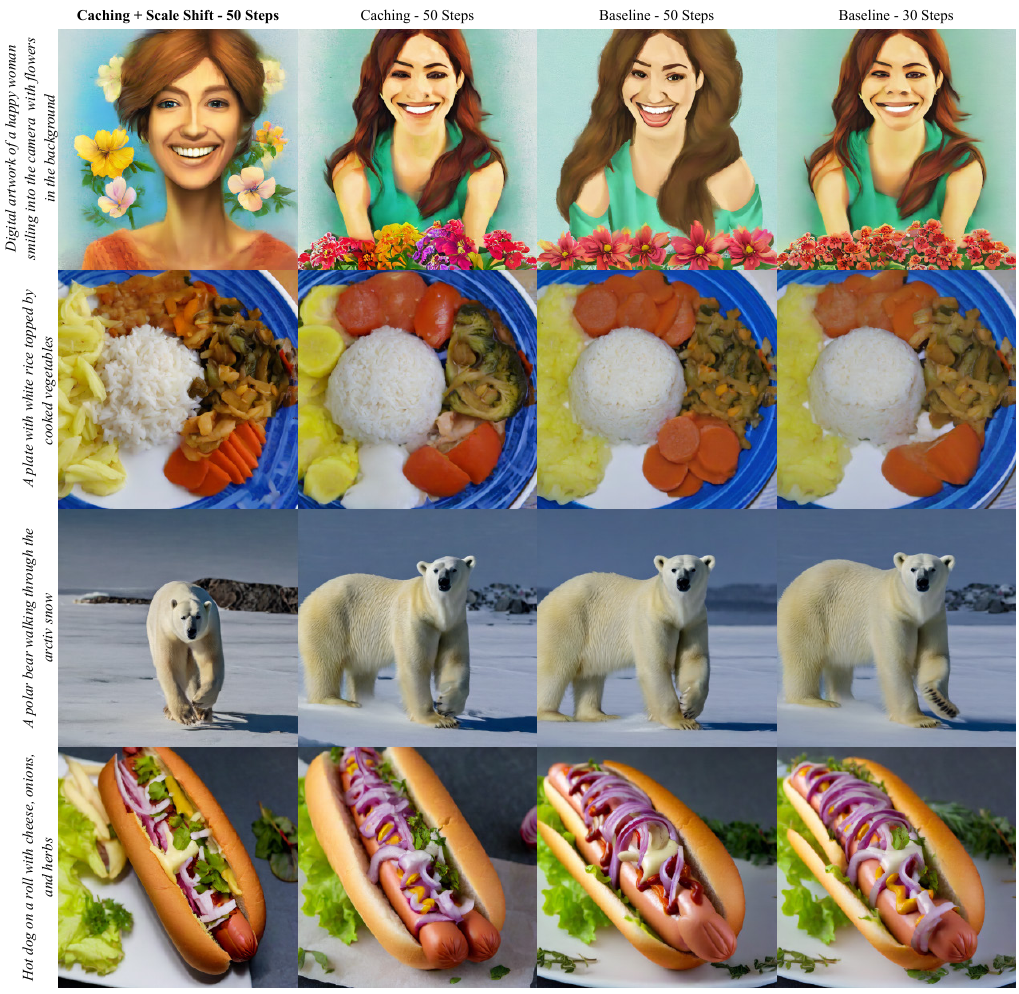}
\caption{
\textbf{Qualitative Results for LDM-512 - DDIM 50 Steps.}}
\label{fig:supp_ldm_ddim50}
\end{figure*}

%% file: figures_supp/change_metrics.tex
\begin{figure}
\centering
\includegraphics[trim={0cm 0cm 0cm 0cm},clip,width=\linewidth]{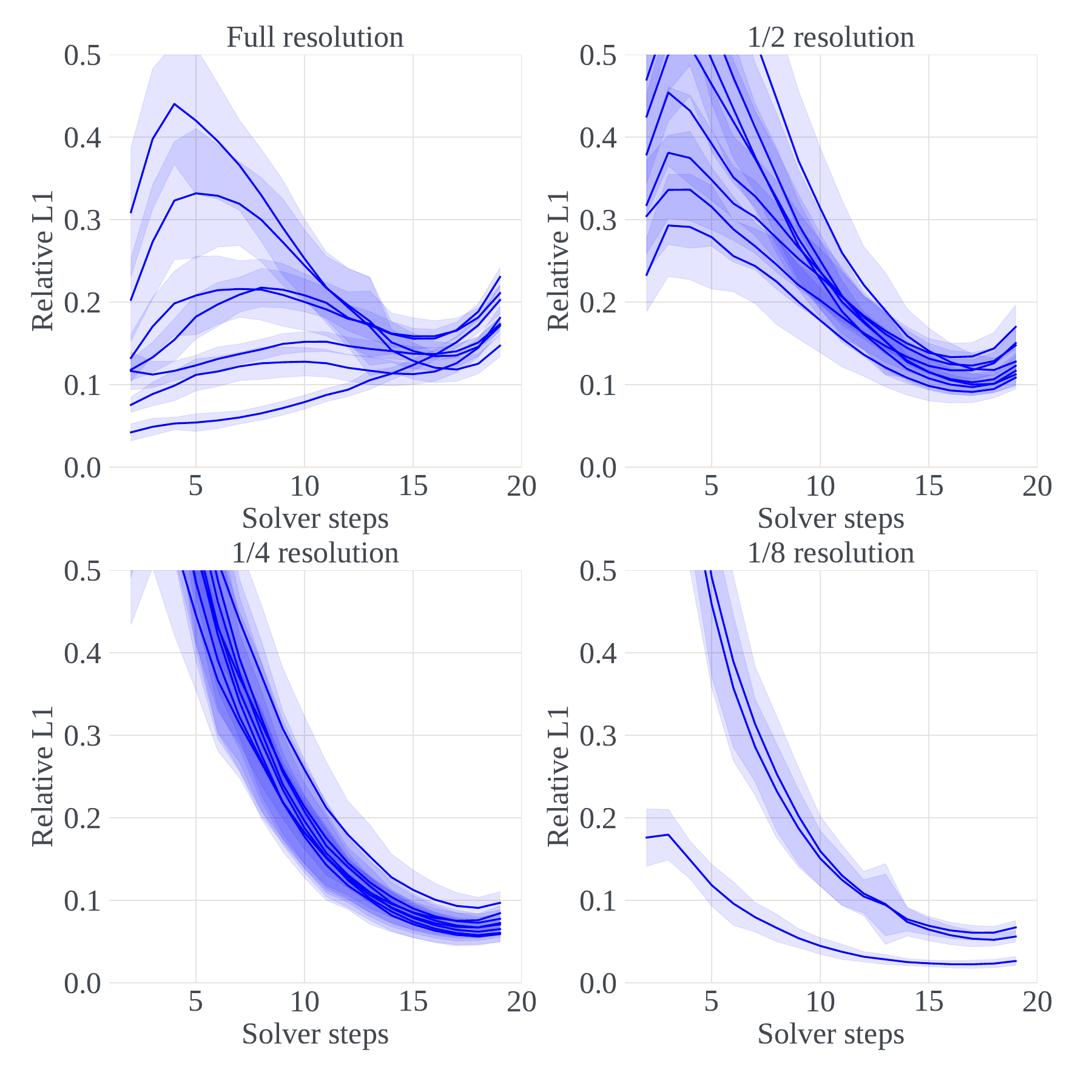}
\caption{
\textbf{Change Metrics for EMU-768 - DPM 20 Steps.}}
\label{fig:supp_change_emu_dpm20}
\end{figure}

\begin{figure}
\centering
\includegraphics[trim={0cm 0cm 0cm 0cm},clip,width=\linewidth]{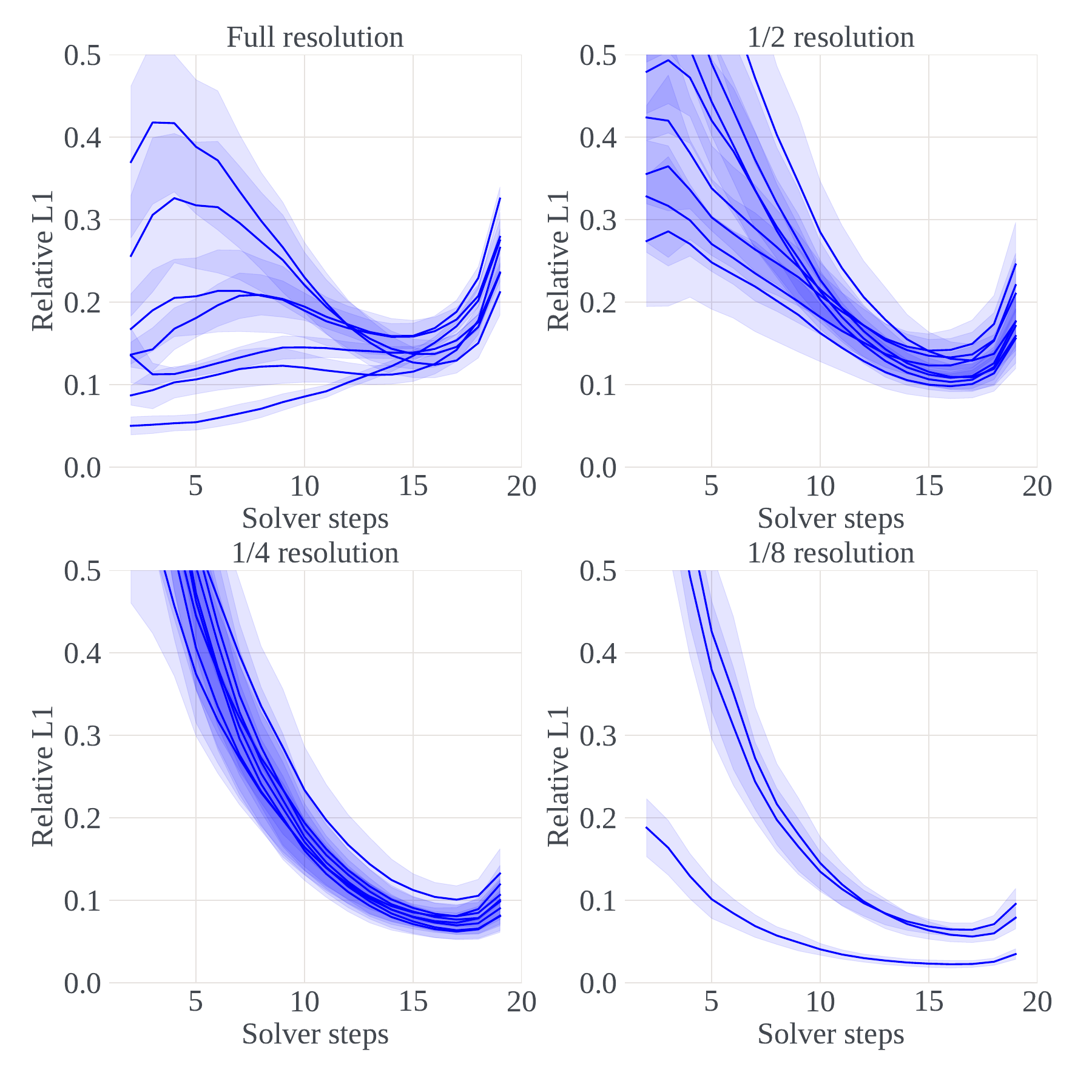}
\caption{
\textbf{Change Metrics for EMU-768 - DDIM 20 Steps.}}
\label{fig:supp_change_emu_ddim20}
\end{figure}

\begin{figure}
\centering
\includegraphics[trim={0cm 0cm 0cm 0cm},clip,width=\linewidth]{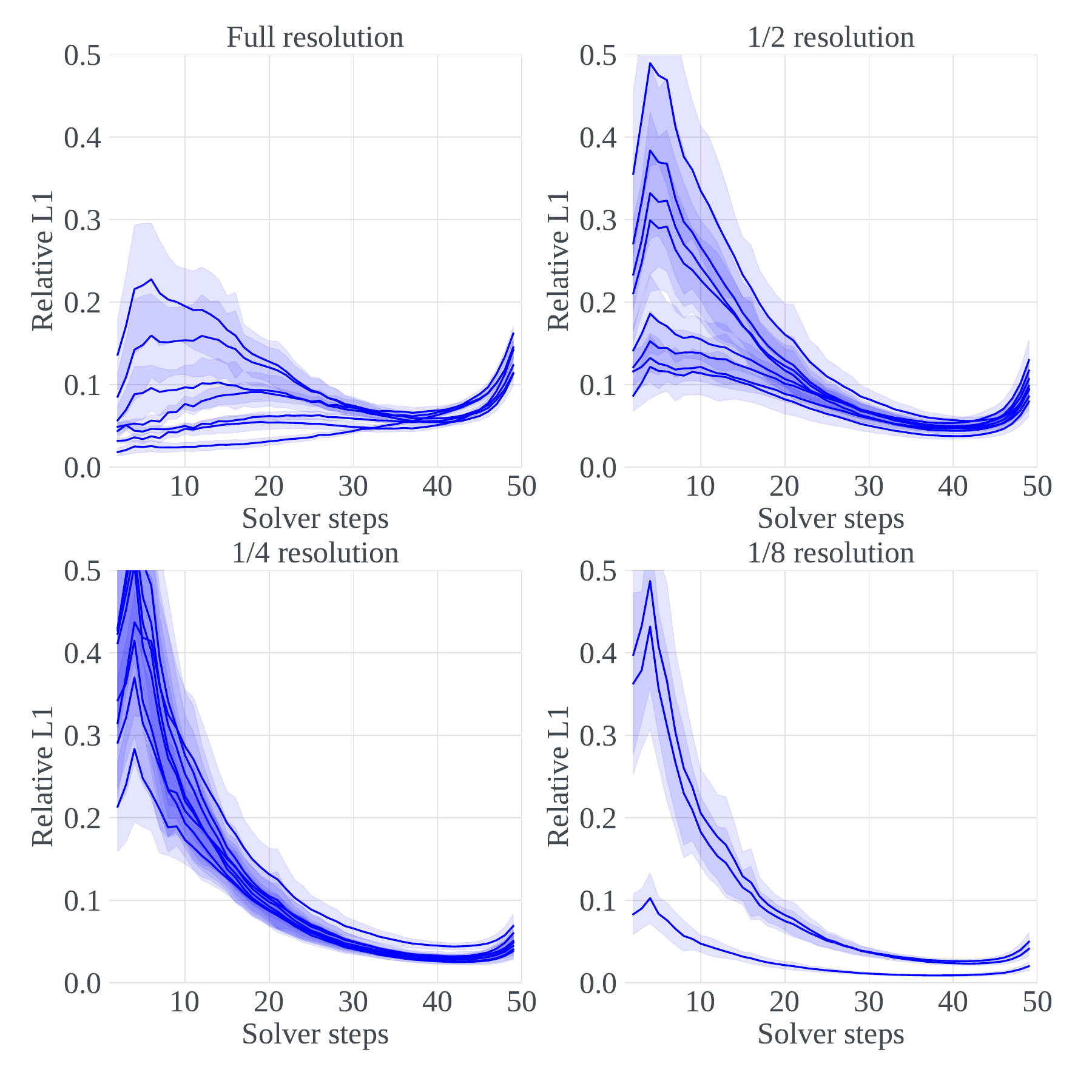}
\caption{
\textbf{Change Metrics for EMU-768 - DPM 50 Steps.}}
\label{fig:supp_change_emu_dpm50}
\end{figure}

\begin{figure}
\centering
\includegraphics[trim={0cm 0cm 0cm 0cm},clip,width=\linewidth]{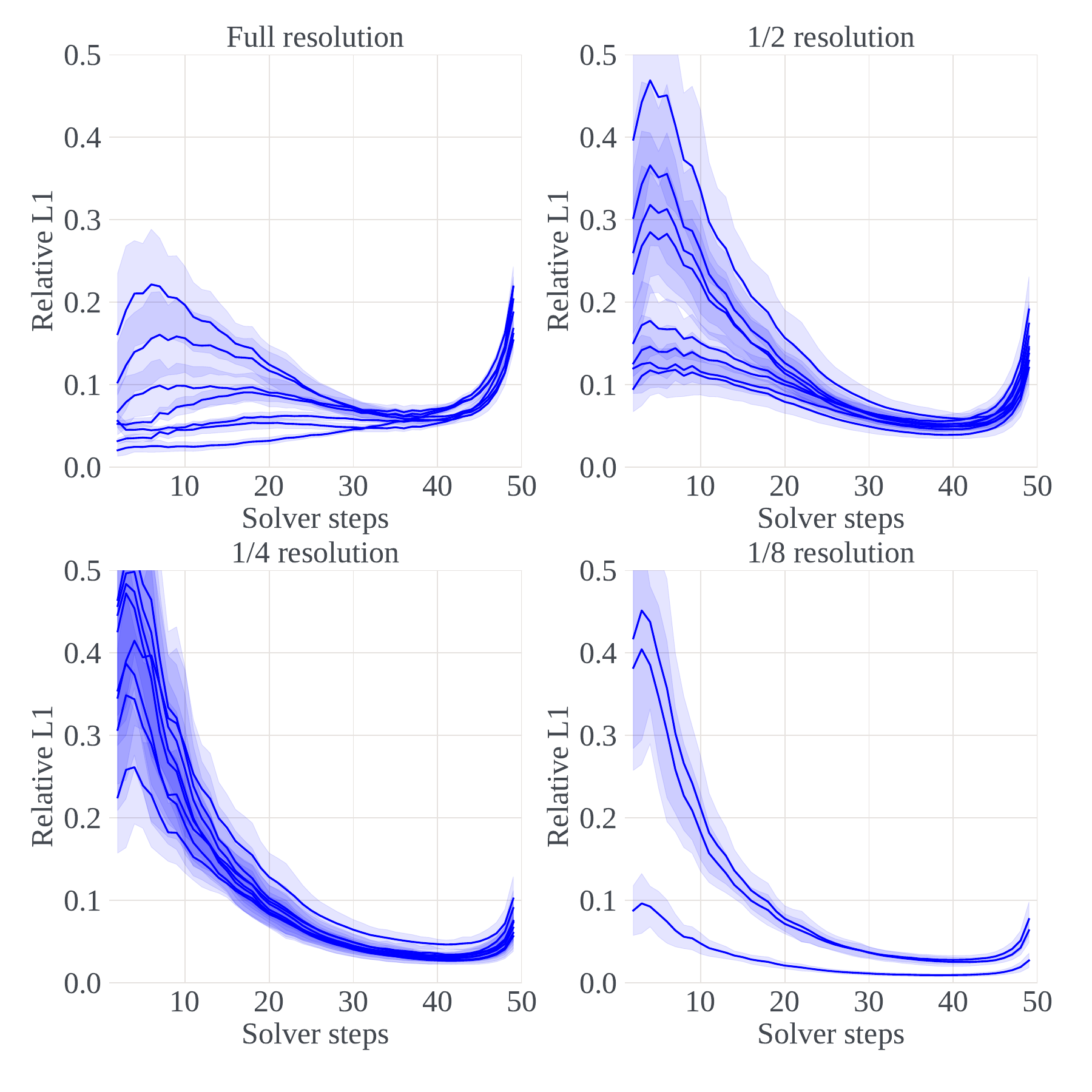}
\caption{
\textbf{Change Metrics for EMU-768 - DDIM 50 Steps.}}
\label{fig:supp_change_emu_ddim50}
\end{figure}

\begin{figure}
\centering
\includegraphics[trim={0cm 0cm 0cm 0cm},clip,width=\linewidth]{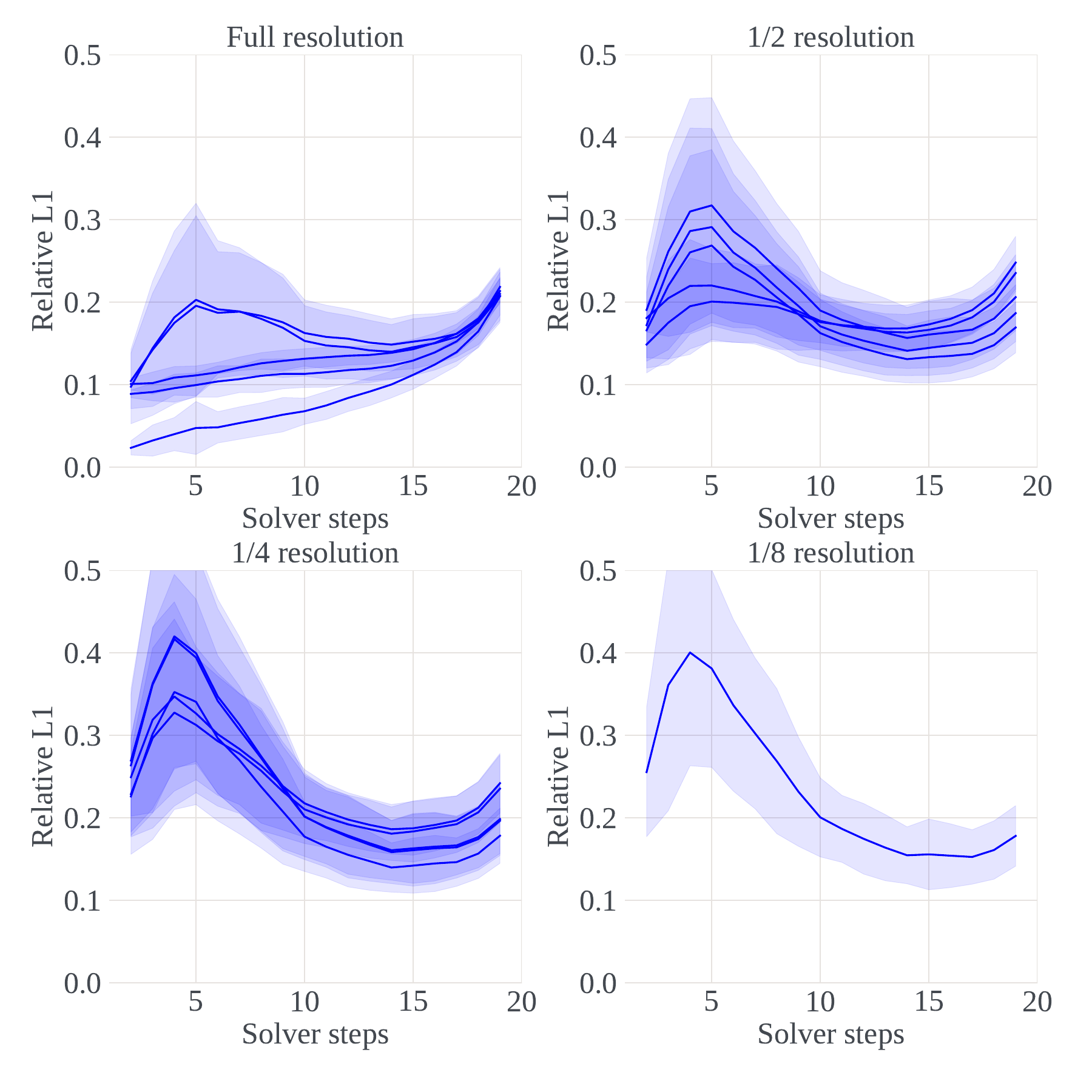}
\caption{
\textbf{Change Metrics for LDM-512 - DPM 20 Steps.}}
\label{fig:supp_change_ldm_dpm20}
\end{figure}

\begin{figure}
\centering
\includegraphics[trim={0cm 0cm 0cm 0cm},clip,width=\linewidth]{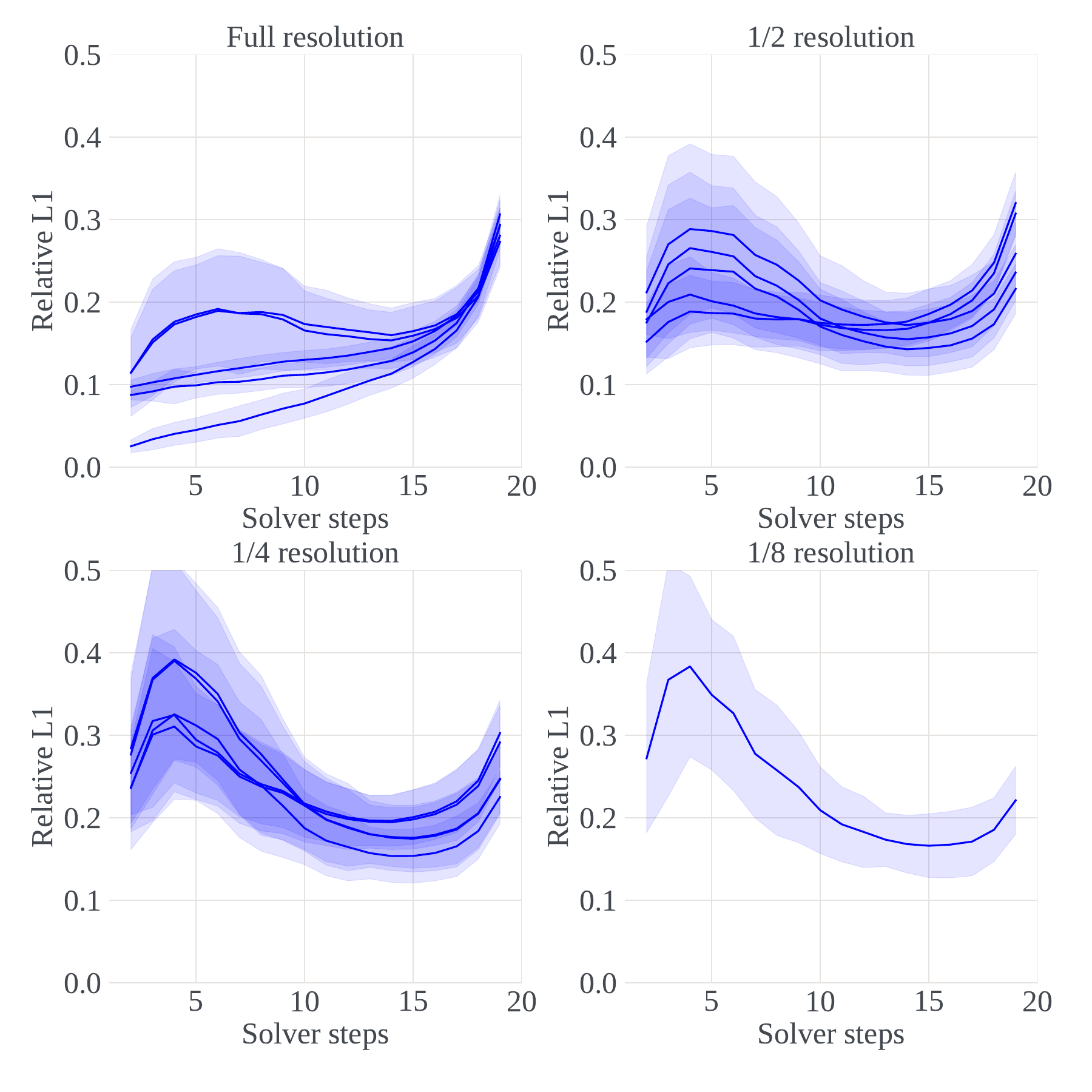}
\caption{
\textbf{Change Metrics for LDM-512 - DDIM 20 Steps.}}
\label{fig:supp_change_ldm_ddim20}
\end{figure}

\begin{figure}
\centering
\includegraphics[trim={0cm 0cm 0cm 0cm},clip,width=\linewidth]{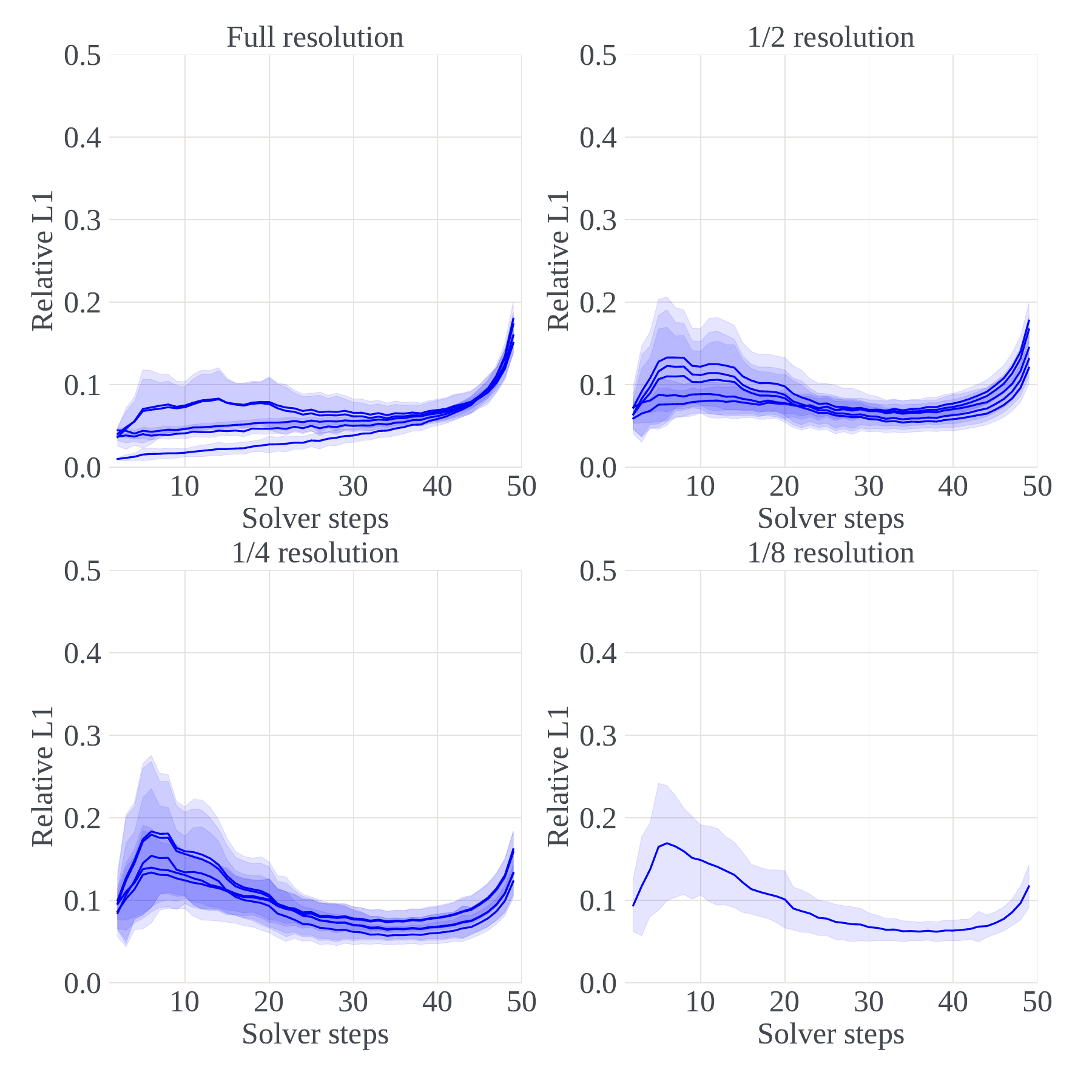}
\caption{
\textbf{Change Metrics for LDM-512 - DPM 50 Steps.}}
\label{fig:supp_change_ldm_dpm50}
\end{figure}

\begin{figure}
\centering
\includegraphics[trim={0cm 0cm 0cm 0cm},clip,width=\linewidth]{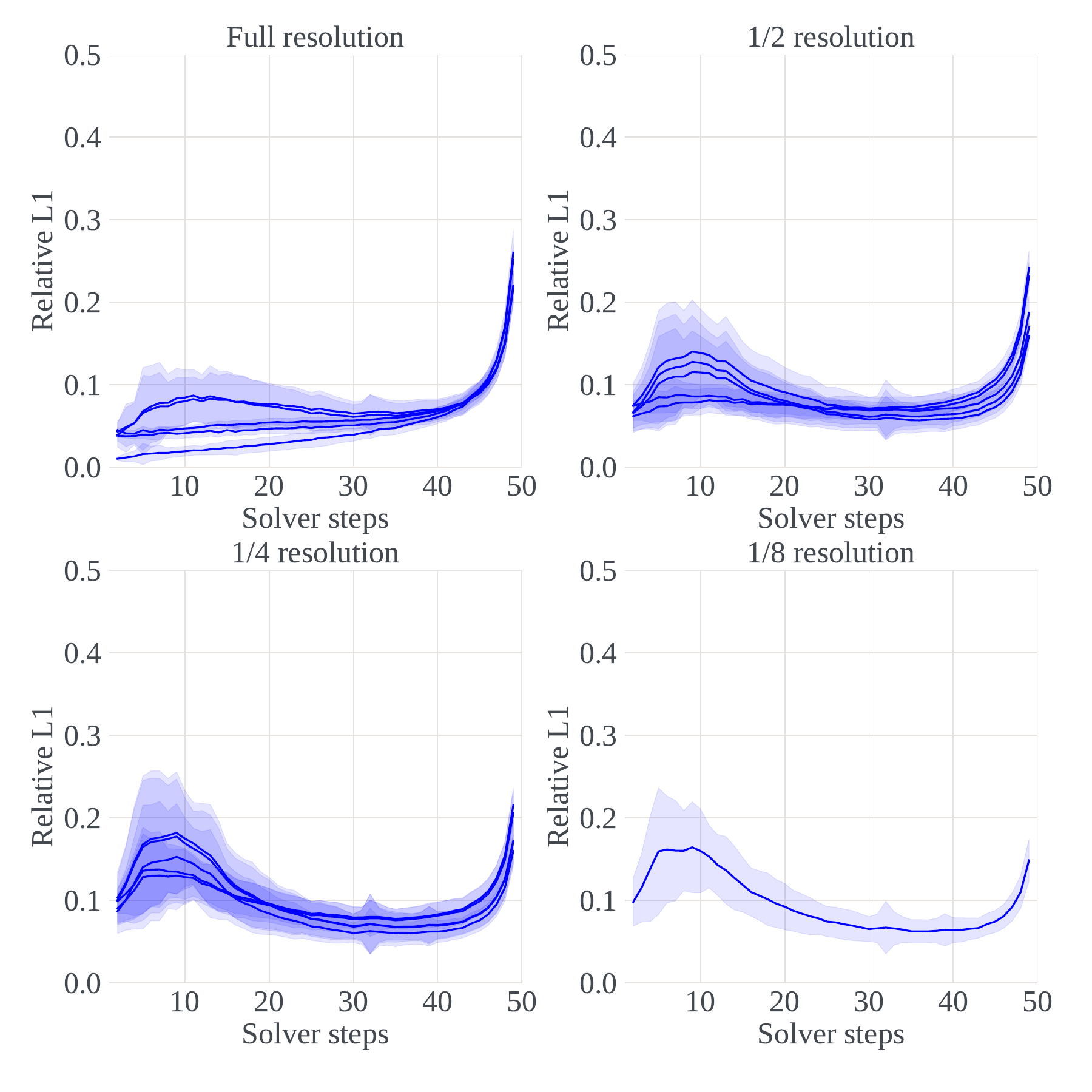}
\caption{
\textbf{Change Metrics for LDM-512 - DDIM 50 Steps.}}
\label{fig:supp_change_ldm_ddim50}
\end{figure}

%% file: figures_supp/cache_schedules.tex
\begin{figure}
\centering
\includegraphics[trim={0cm 0cm 0cm 0cm},clip,width=\linewidth]{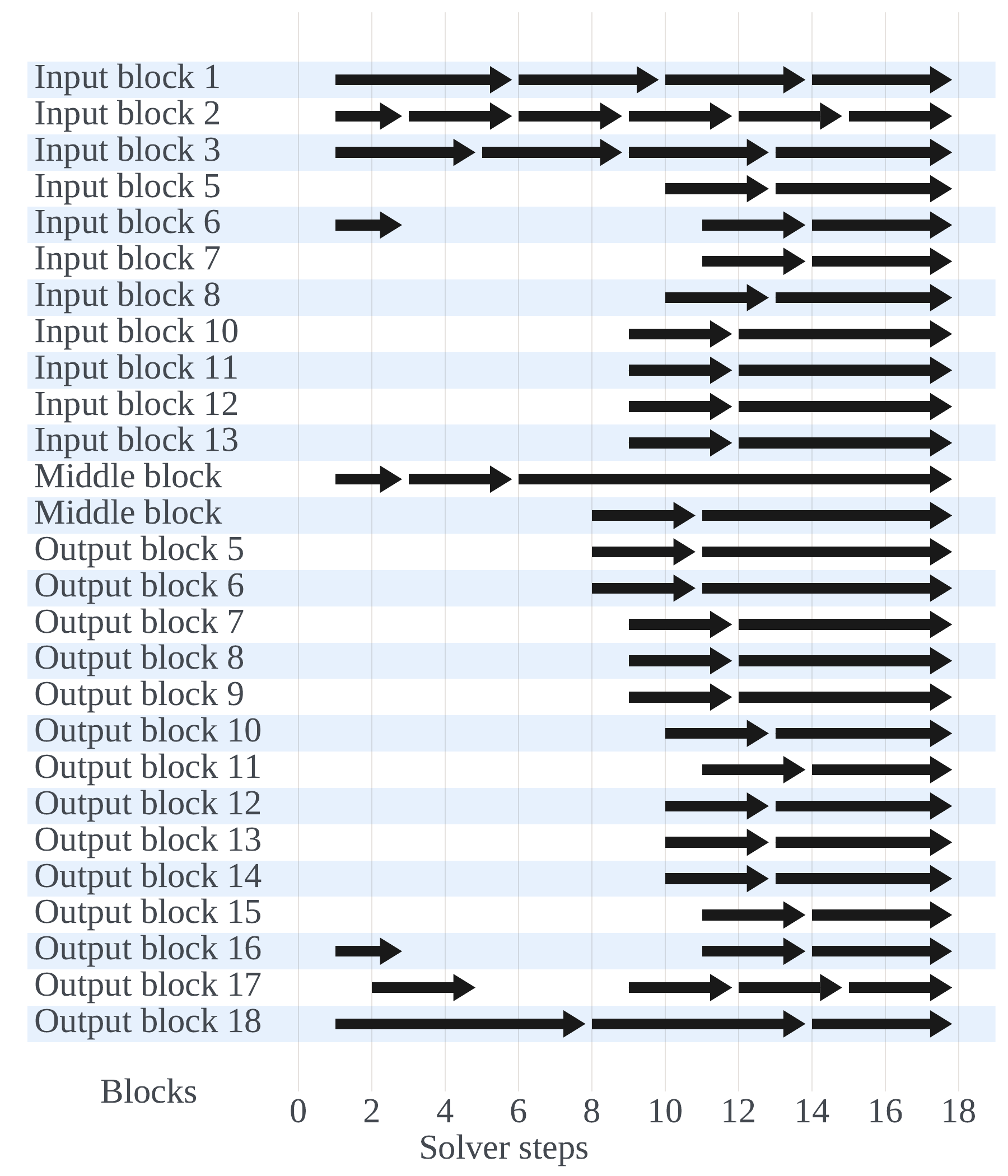}
\caption{
\textbf{Cache Schedules for EMU-768 - DPM 20 Steps.}}
\label{fig:supp_schedule_emu_dpm20}
\end{figure}

\begin{figure}
\centering
\includegraphics[trim={0cm 0cm 0cm 0cm},clip,width=\linewidth]{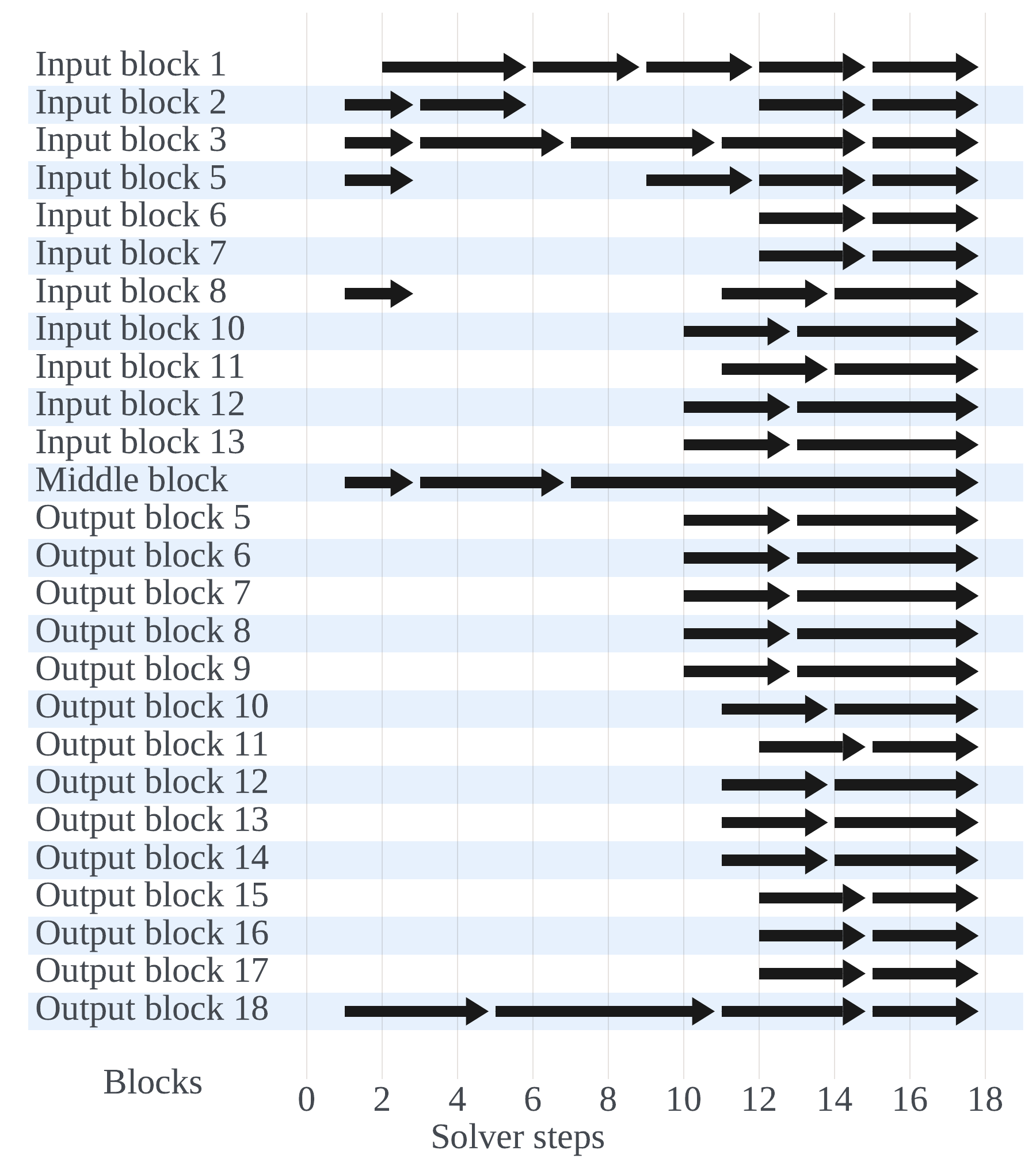}
\caption{
\textbf{Cache Schedules for EMU-768 - DDIM 20 Steps.}}
\label{fig:supp_schedule_emu_ddim20}
\end{figure}

\begin{figure}
\centering
\includegraphics[trim={0cm 0cm 0cm 0cm},clip,width=\linewidth]{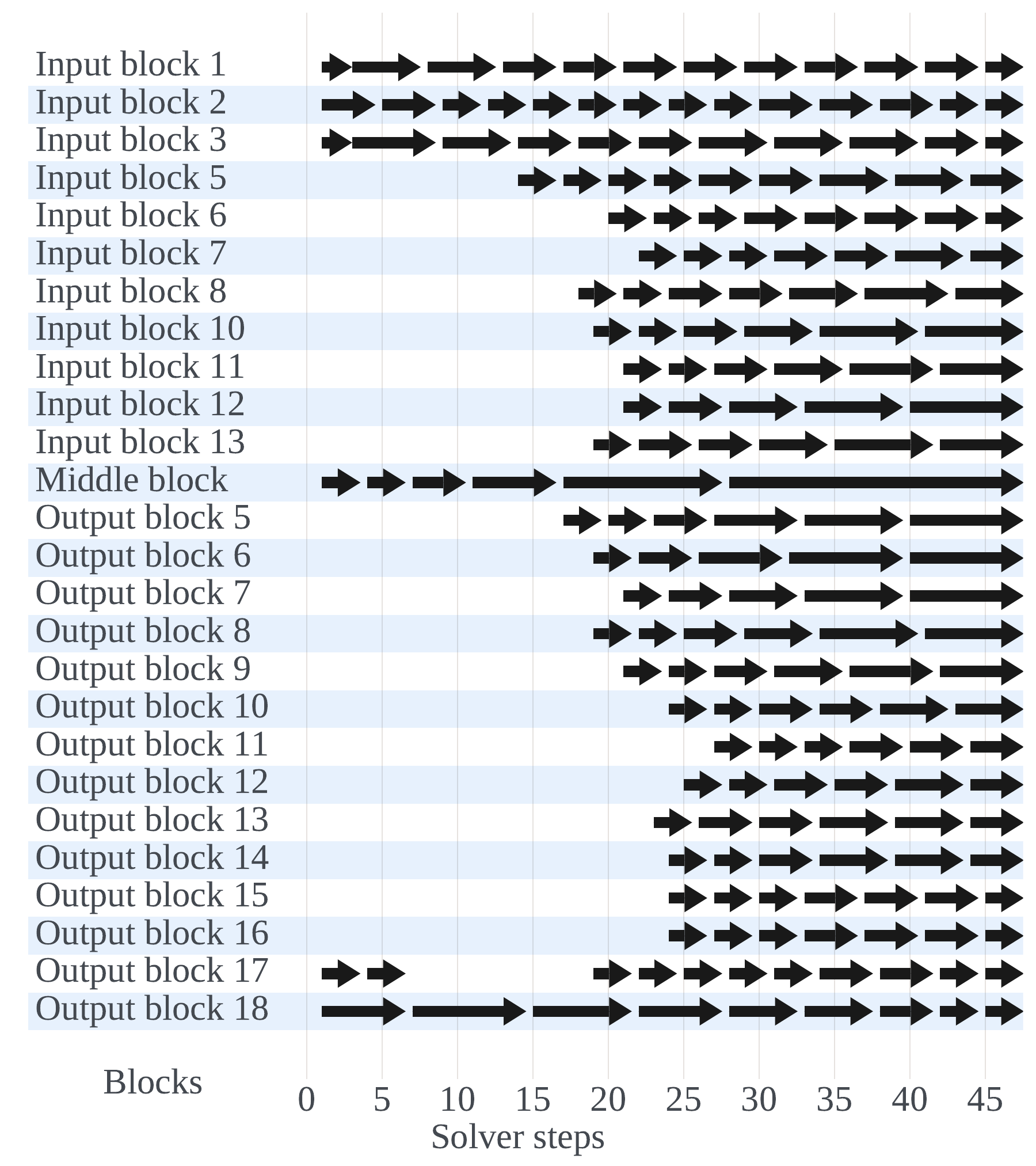}
\caption{
\textbf{Cache Schedules for EMU-768 - DPM 50 Steps.}}
\label{fig:supp_schedule_emu_dpm50}
\end{figure}

\begin{figure}
\centering
\includegraphics[trim={0cm 0cm 0cm 0cm},clip,width=\linewidth]{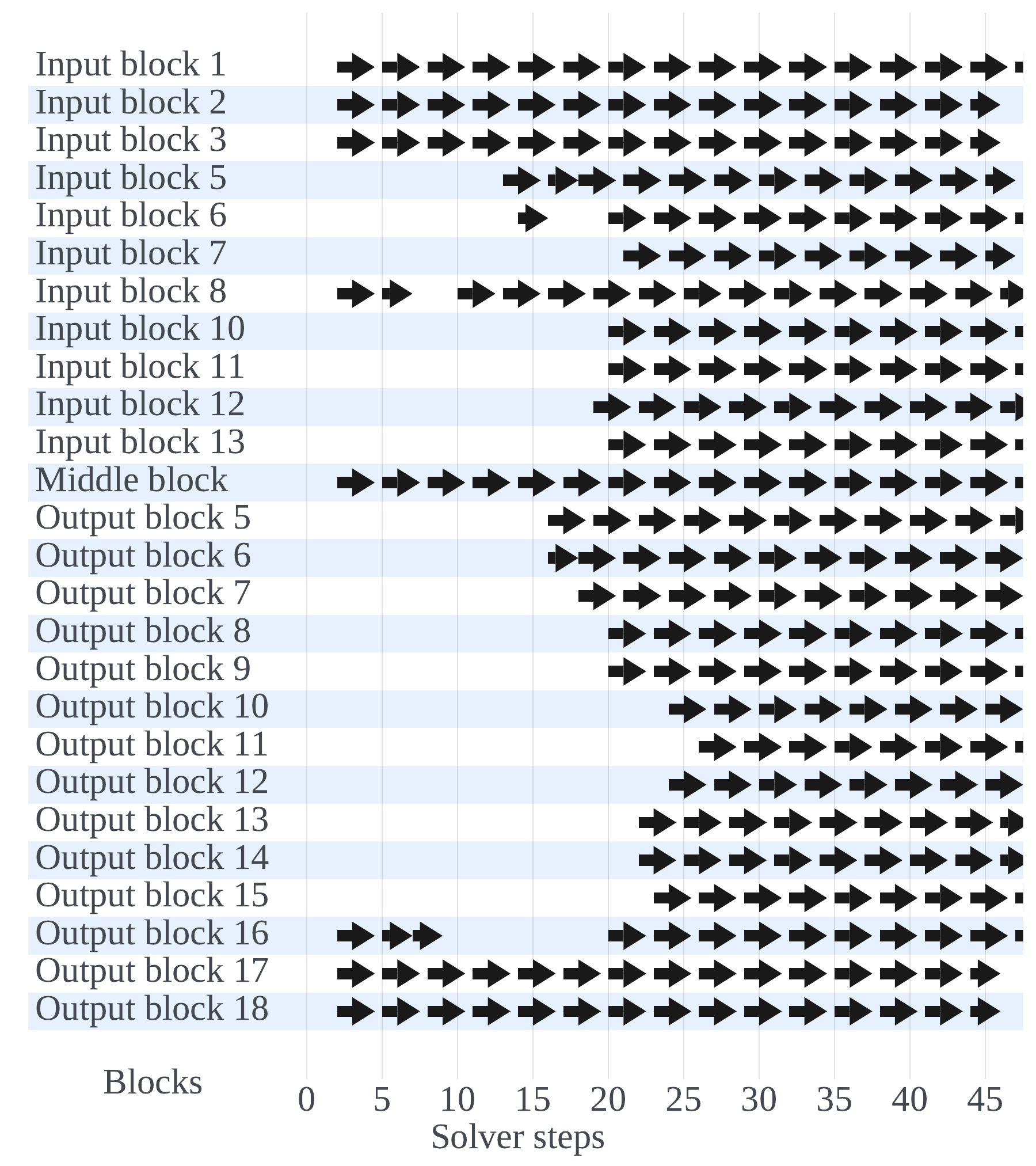}
\caption{
\textbf{Cache Schedules for EMU-768 - DDIM 50 Steps.}}
\label{fig:supp_schedule_emu_ddim50}
\end{figure}

\begin{figure}
\centering
\includegraphics[trim={0cm 0cm 0cm 0cm},clip,width=\linewidth]{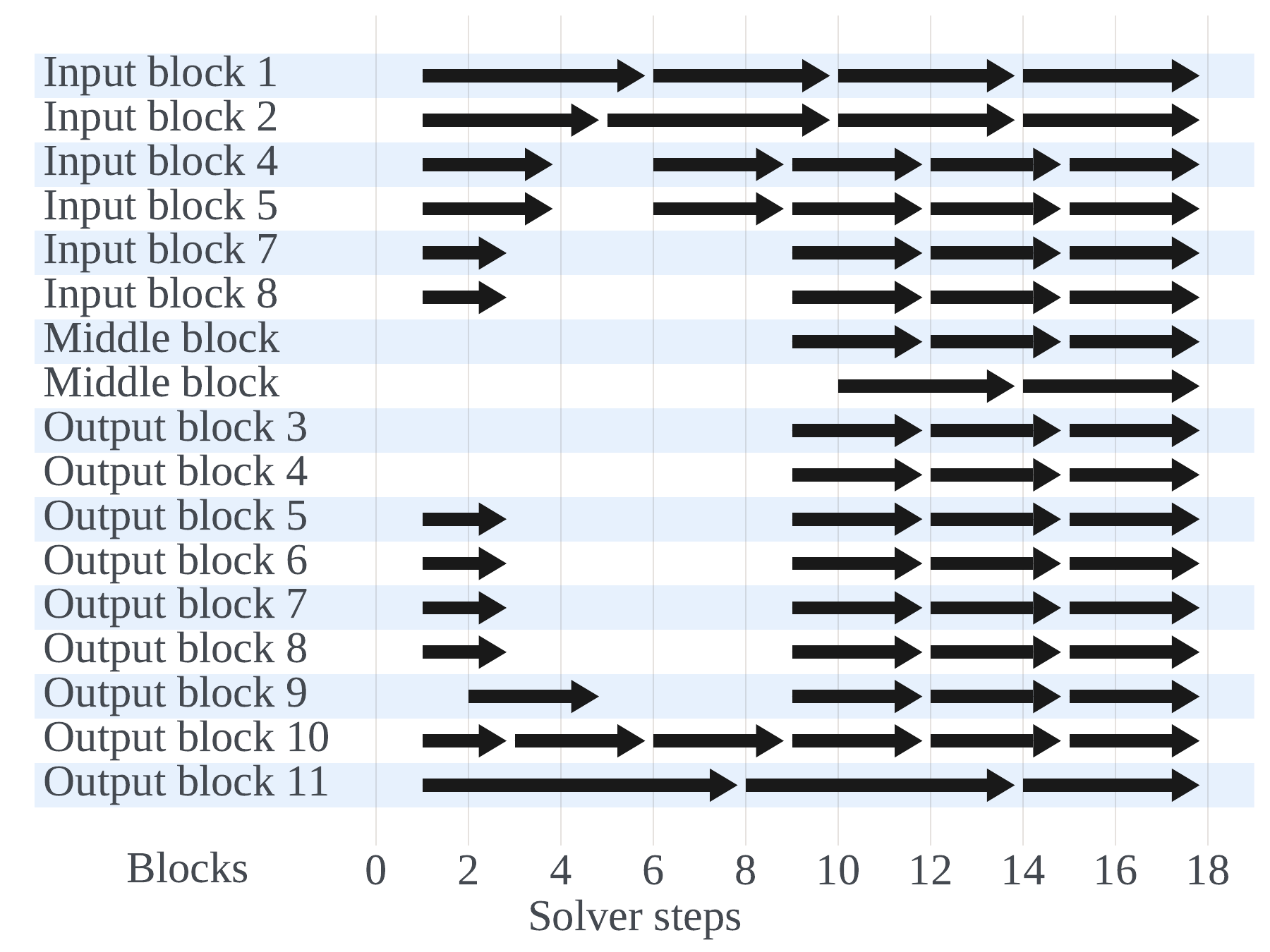}
\caption{
\textbf{Cache Schedules for LDM-512 - DPM 20 Steps.}}
\label{fig:supp_schedule_ldm_dpm20}
\vspace{5cm}
\end{figure}

\begin{figure}
\centering
\includegraphics[trim={0cm 0cm 0cm 0cm},clip,width=\linewidth]{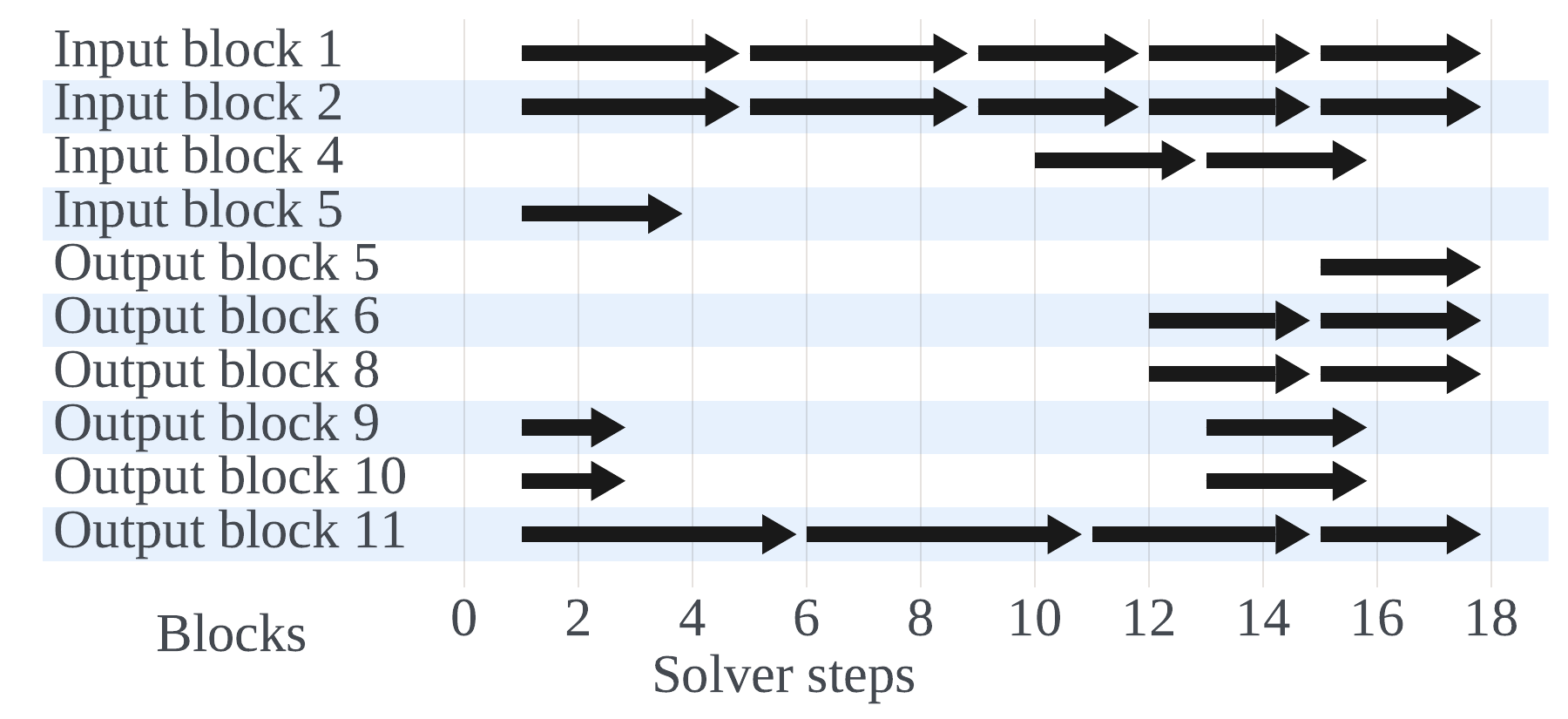}
\caption{
\textbf{Cache Schedules for LDM-512 - DDIM 20 Steps.}}
\label{fig:supp_schedule_ldm_ddim20}
\end{figure}

\begin{figure}[t]
\vspace{.75cm}
\centering
\includegraphics[trim={0cm 0cm 0cm 0cm},clip,width=\linewidth]{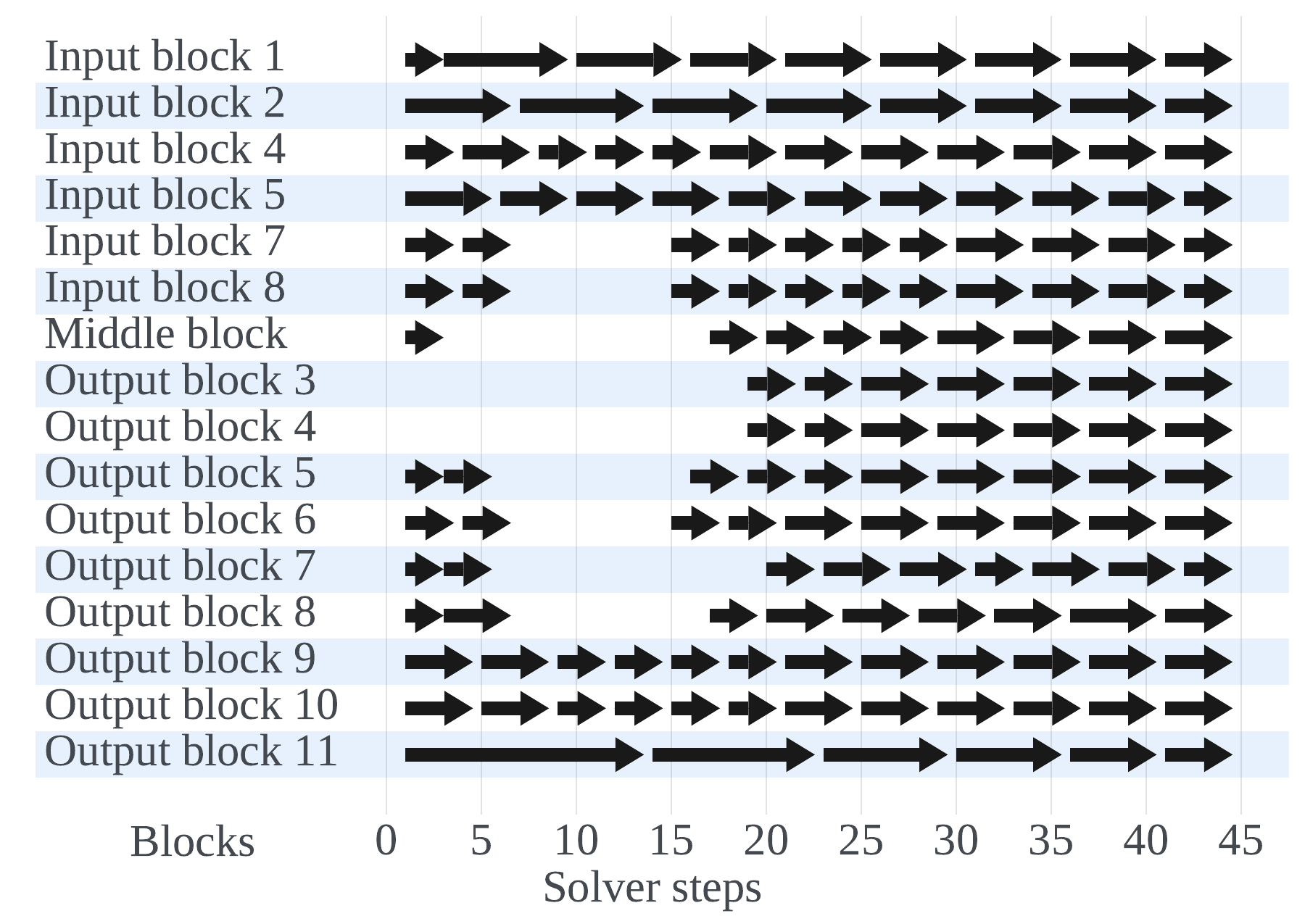}
\caption{
\textbf{Cache Schedules for LDM-512 - DPM 50 Steps.}}
\label{fig:supp_schedule_ldm_dpm50}
\vspace{5cm}
\end{figure}

\begin{figure}

\centering
\includegraphics[trim={0cm 0cm 0cm 0cm},clip,width=\linewidth]{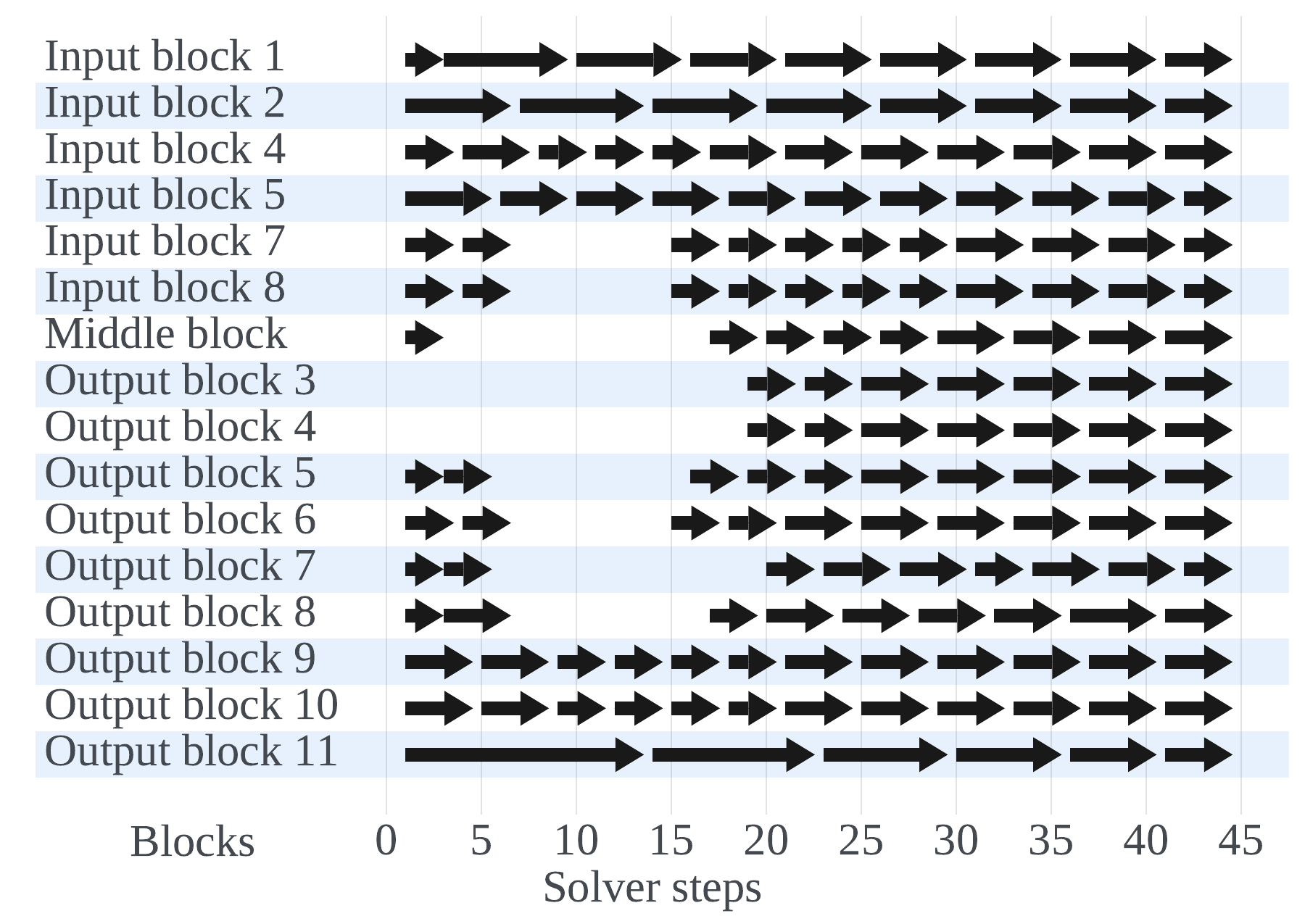}
\caption{
\textbf{Cache Schedules for LDM-512 - DDIM 50 Steps.}}
\label{fig:supp_schedule_ldm_ddim50} 
\end{figure}

\newpage
$ $
\newpage
$ $
\newpage
$ $
\newpage
$ $
\newpage
$ $
\newpage
$ $
\newpage
$ $
\newpage
$ $
\newpage
$ $
\newpage
$ $
\newpage
$ $
\newpage
$ $
\newpage
$ $
\newpage
$ $
\newpage